\documentclass{article}
\usepackage{ijcai24}

\usepackage{times}
\usepackage{soul}
\usepackage{url}
\usepackage[hidelinks]{hyperref}
\usepackage[utf8]{inputenc}
\usepackage[small]{caption}
\usepackage{graphicx}
\usepackage{amsthm}
\usepackage{booktabs}
\usepackage{algorithm}
\usepackage[noEnd=true]{algpseudocodex} 
\usepackage[switch]{lineno}

\usepackage[textsize=scriptsize,colorinlistoftodos]{todonotes}
\usepackage{mathtools}
\usepackage{stmaryrd}
\usepackage{amsmath}
\usepackage[rgb]{optional}
\usepackage[inline]{enumitem}
\usepackage{listings}
\usepackage[mode=buildmissing]{standalone}
\usepackage{ifthen}
\usepackage{amssymb}
\usepackage[frozencache]{minted}
\usepackage{placeins}

\usepackage[createShortEnv,conf={restate, no link to proof,text proof={Proof}}]{proof-at-the-end}

\newboolean{techreport}
\setboolean{techreport}{true}

\opt{cmyk}{
\definecolor{rwth-blue}{cmyk}{1,.5,0,0}
\definecolor{rwth-lblue}{cmyk}{0.75,0.38,0,0}
\definecolor{rwth-llblue}{cmyk}{0.45,0.14,0,0}
\definecolor{rwth-lllblue}{cmyk}{0.23,0.07,0,0}
\definecolor{rwth-llllblue}{cmyk}{0.09,0.03,0,0}

\definecolor{rwth-black}{cmyk}{0,0,0,1}
\colorlet{rwth-lblack}{rwth-black!75}
\colorlet{rwth-llblack}{rwth-black!50}
\colorlet{rwth-lllblack}{rwth-black!25}
\colorlet{rwth-llllblack}{rwth-black!10}

\definecolor{rwth-magenta}{cmyk}{0,1,.25,0}
\colorlet{rwth-lmagenta}{rwth-magenta!75}
\colorlet{rwth-llmagenta}{rwth-magenta!50}
\colorlet{rwth-lllmagenta}{rwth-magenta!25}
\colorlet{rwth-llllmagenta}{rwth-magenta!10}

\definecolor{rwth-yellow}{cmyk}{0,0,1,0}
\colorlet{rwth-lyellow}{rwth-yellow!75}
\colorlet{rwth-llyellow}{rwth-yellow!50}
\colorlet{rwth-lllyellow}{rwth-yellow!25}
\colorlet{rwth-llllyellow}{rwth-yellow!10}

\definecolor{rwth-petrol}{cmyk}{1,0.3,0.5,0.3}
\colorlet{rwth-lpetrol}{rwth-petrol!75}
\colorlet{rwth-llpetrol}{rwth-petrol!50}
\colorlet{rwth-lllpetrol}{rwth-petrol!25}
\colorlet{rwth-llllpetrol}{rwth-petrol!10}

\definecolor{rwth-turquoise}{cmyk}{1,0,0.4,0}
\colorlet{rwth-lturquoise}{rwth-turquoise!75}
\colorlet{rwth-llturquoise}{rwth-turquoise!50}
\colorlet{rwth-lllturquoise}{rwth-turquoise!25}
\colorlet{rwth-llllturquoise}{rwth-turquoise!10}

\definecolor{rwth-green}{cmyk}{.7,0,1,0}
\colorlet{rwth-lgreen}{rwth-green!75}
\colorlet{rwth-llgreen}{rwth-green!50}
\colorlet{rwth-lllgreen}{rwth-green!25}
\colorlet{rwth-llllgreen}{rwth-green!10}

\definecolor{rwth-grass}{cmyk}{.35,0,1,0}
\colorlet{rwth-lgrass}{rwth-grass!75}
\colorlet{rwth-llgrass}{rwth-grass!50}
\colorlet{rwth-lllgrass}{rwth-grass!25}
\colorlet{rwth-llllgrass}{rwth-grass!10}

\definecolor{rwth-orange}{cmyk}{0,.4,1,0}
\colorlet{rwth-lorange}{rwth-orange!75}
\colorlet{rwth-llorange}{rwth-orange!50}
\colorlet{rwth-lllorange}{rwth-orange!25}
\colorlet{rwth-llllorange}{rwth-orange!10}

\definecolor{rwth-red}{cmyk}{.15,1,1,0}
\colorlet{rwth-lred}{rwth-red!75}
\colorlet{rwth-llred}{rwth-red!50}
\colorlet{rwth-lllred}{rwth-red!25}
\colorlet{rwth-llllred}{rwth-red!10}

\definecolor{rwth-burgundy}{cmyk}{0.25,1,0.7,0.2}
\colorlet{rwth-lburgundy}{rwth-burgundy!75}
\colorlet{rwth-llburgundy}{rwth-burgundy!50}
\colorlet{rwth-lllburgundy}{rwth-burgundy!25}
\colorlet{rwth-llllburgundy}{rwth-burgundy!10}

\definecolor{rwth-violet}{cmyk}{0.7,1,0.35,0.15}
\colorlet{rwth-lviolet}{rwth-violet!75}
\colorlet{rwth-llviolet}{rwth-violet!50}
\colorlet{rwth-lllviolet}{rwth-violet!25}
\colorlet{rwth-llllviolet}{rwth-violet!10}

\definecolor{rwth-purple}{cmyk}{0.6,0.6,0,0}
\colorlet{rwth-lpurple}{rwth-purple!75}
\colorlet{rwth-llpurple}{rwth-purple!50}
\colorlet{rwth-lllpurple}{rwth-purple!25}
\colorlet{rwth-llllpurple}{rwth-purple!10}

\definecolor{rwth-cyan}{cmyk}{1,0,.4,0}\colorlet{rwth-lcyan}{rwth-cyan!50}\colorlet{rwth-llcyan}{rwth-cyan!25}
\definecolor{rwth-teal}{cmyk}{1,.3,.5,.3}\colorlet{rwth-lteal}{rwth-teal!50}\colorlet{rwth-llteal}{rwth-teal!25}
\definecolor{rwth-silver}{cmyk}{.39,.31,.32,.14}
\definecolor{rwth-gold}{cmyk}{.35,.46,.7,.35}
}

\opt{rgb}{
\definecolor{rwth-blue}{RGB}{0,84,159}
\definecolor{rwth-lblue}{RGB}{64,127,183}
\definecolor{rwth-llblue}{RGB}{142,186,229}
\definecolor{rwth-lllblue}{RGB}{199,221,242}
\definecolor{rwth-llllblue}{RGB}{232,241,250}

\definecolor{rwth-black}{RGB}{0,0,0}
\definecolor{rwth-lblack}{RGB}{100,101,103}
\definecolor{rwth-llblack}{RGB}{156,158,159}
\definecolor{rwth-lllblack}{RGB}{208,209,210}
\definecolor{rwth-llllblack}{RGB}{236,237,237}

\definecolor{rwth-magenta}{RGB}{227,0,102}
\definecolor{rwth-lmagenta}{RGB}{233,96,136}
\definecolor{rwth-llmagenta}{RGB}{241,158,177}
\definecolor{rwth-lllmagenta}{RGB}{249,210,218}
\definecolor{rwth-llllmagenta}{RGB}{253,238,240}

\definecolor{rwth-yellow}{RGB}{255,237,0}
\definecolor{rwth-lyellow}{RGB}{255,240,85}
\definecolor{rwth-llyellow}{RGB}{255,245,155}
\definecolor{rwth-lllyellow}{RGB}{255,250,209}
\definecolor{rwth-llllyellow}{RGB}{255,253,,238}

\definecolor{rwth-petrol}{RGB}{0,97,101}
\definecolor{rwth-lpetrol}{RGB}{45,127,131}
\definecolor{rwth-llpetrol}{RGB}{125,164,167}
\definecolor{rwth-lllpetrol}{RGB}{191,208,209}
\definecolor{rwth-llllpetrol}{RGB}{230,236,236}

\definecolor{rwth-turquoise}{RGB}{0,152,161}
\definecolor{rwth-lturquoise}{RGB}{0,177,183}
\definecolor{rwth-llturquoise}{RGB}{137,204,207}
\definecolor{rwth-lllturquoise}{RGB}{202,231,231}
\definecolor{rwth-llllturquoise}{RGB}{235,246,246}

\definecolor{rwth-green}{RGB}{87,171,39}
\definecolor{rwth-lgreen}{RGB}{141,192,96}
\definecolor{rwth-llgreen}{RGB}{184,214,152}
\definecolor{rwth-lllgreen}{RGB}{221,235,206}
\definecolor{rwth-llllgreen}{RGB}{242,247,236}

\definecolor{rwth-grass}{RGB}{189,205,0}
\definecolor{rwth-lgrass}{RGB}{208,217,92}
\definecolor{rwth-llgrass}{RGB}{224,230,154}
\definecolor{rwth-lllgrass}{RGB}{240,43,208}
\definecolor{rwth-llllgrass}{RGB}{249,250,237}

\definecolor{rwth-orange}{RGB}{246,168,0}
\definecolor{rwth-lorange}{RGB}{250,190,80}
\definecolor{rwth-llorange}{RGB}{253,212,143}
\definecolor{rwth-lllorange}{RGB}{254,234,201}
\definecolor{rwth-llllorange}{RGB}{255,247,234}

\definecolor{rwth-red}{RGB}{204,7,30}
\definecolor{rwth-lred}{RGB}{216,92,65}
\definecolor{rwth-llred}{RGB}{230,150,121}
\definecolor{rwth-lllred}{RGB}{243,205,187}
\definecolor{rwth-llllred}{RGB}{250,235,227}

\definecolor{rwth-burgundy}{RGB}{161,16,53}
\definecolor{rwth-lburgundy}{RGB}{182,82,86}
\definecolor{rwth-llburgundy}{RGB}{205,139,135}
\definecolor{rwth-lllburgundy}{RGB}{229,197,192}
\definecolor{rwth-llllburgundy}{RGB}{245,232,229}

\definecolor{rwth-violet}{RGB}{97,33,88}
\definecolor{rwth-lviolet}{RGB}{131,78,117}
\definecolor{rwth-llviolet}{RGB}{168,133,158}
\definecolor{rwth-lllviolet}{RGB}{210,192,205}
\definecolor{rwth-llllviolet}{RGB}{237,229,234}

\definecolor{rwth-purple}{RGB}{122,111,172}
\definecolor{rwth-lpurple}{RGB}{122,111,172}
\definecolor{rwth-llpurple}{RGB}{122,111,172}
\definecolor{rwth-lllpurple}{RGB}{122,111,172}
\definecolor{rwth-llllpurple}{RGB}{122,111,172}

\definecolor{rwth-cyan}{RGB}{0,152,161}
\definecolor{rwth-lcyan}{RGB}{0,177,183}
\definecolor{rwth-llcyan}{RGB}{137,204,207}
\definecolor{rwth-lllcyan}{RGB}{202,231,231}
\definecolor{rwth-llllcyan}{RGB}{235,246,246}

\definecolor{rwth-silver}{cmyk}{.39,.31,.32,.14}
\definecolor{rwth-gold}{cmyk}{.35,.46,.7,.35}
}

\setlist[enumerate,1]{label=(\arabic*),ref=\arabic*}

\hypersetup{
  pdftitle ={Learning Generalized Policies for Fully Observable Non-Deterministic Planning Domains},
  pdfauthor={Till Hofmann, Hector Geffner},
  pdfkeywords={learning in planning and scheduling, planning under uncertainty},
  pdfsubject={IJCAI 2024}
}

\usepackage{amsmath}
\usepackage{xspace}
\newcommand*{\dec}[1]{\ensuremath{{#1\negmedspace\downarrow}}}
\newcommand*{\inc}[1]{\ensuremath{{#1\negmedspace\uparrow}}}
\newcommand*{\mi}[1]{\ensuremath{\mathit{#1}}\xspace}

\DeclareMathOperator*{\FondPrune}{DeadFree}
\DeclareMathOperator*{\alive}{S_A}
\DeclareMathOperator*{\goal}{S_G}
\DeclareMathOperator*{\dead}{S_D}
\DeclareMathOperator*{\solvable}{S_S}
\DeclareMathOperator*{\dist}{dist}
\DeclareMathOperator*{\rdist}{rdist}
\DeclareMathOperator*{\sumdist}{sdist}
\DeclareMathOperator*{\sumrdist}{srdist}
\DeclareMathOperator*{\exone}{Exactly-1}
\newcommand*{\la}{\ensuremath{\langle}}
\newcommand*{\ra}{\ensuremath{\rangle}}

\newtheorem{definition}{Definition}

\newEndThm[all end]{definitionE}{definition}

\newcommand*{\Continue}{\State \textbf{continue}}

\title{Learning Generalized Policies for\\Fully Observable Non-Deterministic Planning Domains }

 \author{ Till Hofmann \And Hector Geffner\\ \affiliations RWTH Aachen University\\ \emails till.hofmann@ml.rwth-aachen.de, hector.geffner@ml.rwth-aachen.de }

\newcommand{\hector}[1]{}
\newcommand{\till}[1]{}

\newcommand{\Q}{\mathcal{Q}}
\newcommand{\Omit}[1]{}
\newcommand{\tup}[1]{\langle #1 \rangle}
\newcommand{\pplus}{\hspace{-.05em}\raisebox{.15ex}{\footnotesize$\uparrow$}}
\newcommand{\mminus}{\hspace{-.05em}\raisebox{.15ex}{\footnotesize$\downarrow$}}

\newcommand{\EQ}[1]{#1{\,{=}\,}0}
\newcommand{\GT}[1]{#1{\,{>}\,}0}
\newcommand{\DEC}[1]{#1\mminus}
\newcommand{\INC}[1]{#1\pplus}
\newcommand{\UNK}[1]{#1?}

\newcommand{\F}{\mathcal{F}}
\renewcommand{\S}{\mathcal{S}}

\begin{document}

\maketitle

\begin{abstract}
  General policies represent reactive strategies for solving large families of planning problems like the infinite collection of solvable instances from a given domain. Methods for learning such policies from a collection of small training instances have been developed successfully for classical domains. In this work, we extend the formulations and the resulting combinatorial methods for learning general policies over fully observable, non-deterministic (FOND) domains. We also evaluate the resulting approach experimentally over a number of benchmark domains in FOND planning, present the general policies that result in some of these domains, and prove their correctness. The method for learning general policies for FOND planning can actually be seen as an alternative FOND planning method that searches for solutions, not in the given state space but in an abstract space defined by features that must be learned as well.

\end{abstract}

\section{Introduction}%
\label{sec:introduction}

General  policies express reactive   strategies for solving large families of planning problems such as all Blocks world problems 
 \cite{srivastava08learning,hu:generalized,BelleL16,bonet:ijcai2018,sheila:generalized2019,sergio:generalized}.
Methods for learning such policies have been developed successfully for classical domains appealing to either combinatorial or deep learning
approaches \cite{karpas:generalized,bonet:aaai2019,stahlberg-et-al-icaps2022}. While the learning methods do not guarantee that the resulting
general policies are correct and will solve all the problems in the target class, the policies obtained from combinatorial methods
are more transparent and can  be analyzed and shown to be correct  on an individual basis \cite{frances:aaai2021,drexler:icaps2022}.

Methods for learning general policies for Markov Decision Problems (MDPs) have also been  developed \cite{sylvie:asnet,mausam:dl,sid:sokoban,babyAI}, 
in most cases relying on deep learning and deep reinforcement learning (DRL) techniques  \cite{dl-book,sutton:book,drl-book}, but the performance of the
learned policies is evaluated experimentally as their correctness cannot be assessed.

The goal of this work is to extend the combinatorial approaches developed for learning general policies for classical domains to
non-deterministic, fully observable (FOND) domains \cite{cimatti:three-models}. The motivations are twofold.
On the one hand, FOND planning is closely related to both classical and MDP planning.
Indeed,  the FOND planners that scale up best are those relying on classical planners \cite{prp,ff-replan,muisePRPRebootedAdvancing2024},
and  the policies that  reach the goal states of an MDP with probability 1 are precisely the policies that
solve the FOND problem underlying the MDP; i.e., where the possible transitions are the ones that have positive probabilities
\cite{geffner:book,ghallab:book}. This means that FOND models capture the qualitative structure of Goal MDPs,
and that general policies that solve classes of  FOND problems  will  also solve correctly a larger class of Goal MDPs.

On the other hand, while the best FOND planners rely on classical planners, FOND planning is harder,
requiring not just exponential time but exponential space.\footnote{Classical planning is PSPACE-hard \cite{bylander:complexity}, 
while FOND planning is EXP-hard \cite{littman:fond,rintanen:complexity}.} So the formal relation between the two planning tasks
is not so clear. Interestingly, this relation becomes clearer in the generalized setting, where, as we will see,  generalized FOND
planning reduces to generalized \emph{classical}  planning plus FOND dead-end detection. In other words, a general  policy
for a class $\Q$ of FOND  problems can be obtained from a general policy for a class $\Q_D$ of classical problems 
obtained by the outcome relaxation from  those in $\Q$ \cite{ff-replan,prp}, along with a description of the dead-end states
to be avoided. 
The resulting  method for learning general policies for FOND planning can also  be  seen as an alternative
FOND planning method  that solves a FOND problem by solving  a number of classical problems, not in the given state space
but in an abstract space defined by features that must be learned as well.

The rest of the paper is organized as follows. We review related work and background first, and then introduce general FOND policies and a method for learning
them, followed by an evaluation and analysis of the results.

\section{Related Work}%
\label{sec:related-work}

\medskip

\noindent \textbf{General policies for classical domains}.
The problem of learning general policies for classical domains  has a long history \cite{khardon:action,martin:generalized,fern:generalized},
and general policies have been formulated in terms of logic \cite{srivastava:generalized,sheila:generalized2019}, and more recently in
terms of features and rules  \cite{bonet:ijcai2018,bonet:aaai2019} that can be learned using combinatorial methods \cite{frances:aaai2021}.
The rule   language has  also been used to express problem  decompositions or sketches \cite{drexler:icaps2021,drexler:icaps2022},
and in this work it  will be used to express general policies for FOND problems.

\medskip

\noindent \textbf{General policies for MDPs}. Deep learning (DL) and deep reinforcement learning (DRL) methods have also been used to learn general
policies for classical domains \cite{sid:sokoban,babyAI,karpas:generalized,stahlberg-et-al-kr2022,simon:kr2023} and MDPs
\cite{boutilier2001symbolic,wang2008first,van-otterlo:survey,sylvie:asnet,mausam:dl,karpas:generalized,sanner:practicalMDPs}.
DL and DRL methods scale up better than combinatorial methods and do not need to assume an existing pool of features, but
the resulting policies are not transparent and cannot be understood or shown to be correct. 

\medskip

\noindent \textbf{FOND planning}.  FOND planning has  become increasingly important as a way of  solving other types of problems,
including MDPs \cite{fond:mdp,camacho:fond}, problems with extended temporal goals \cite{nir:ijcai2013,camacho:ltl,fond:ltl2}
and generalized planning problems \cite{srivastava:aaai2011,bonet:ijcai2017}. FOND planners rely on different techniques
like OBDDs  \cite{cimatti:three-models,gamer}, SAT \cite{fondsat},  graph search \cite{mynd,grendel,pereira:fond},
and classical planning  algorithms \cite{nd,fip,prp,muisePRPRebootedAdvancing2024}, but problems are solved individually from scratch.

\noindent \textbf{Dead-ends}.  Dead-ends in planning refer to states from which there is no solution.
There has been work in learning to identify  dead-ends in classical planning \cite{lipovetzky2016traps,deadends1}, and in
FOND and MDP planning \cite{deadends:mdps,camacho:fond}. Closer to this work is  the learning of \emph{general}  dead-end representations
\cite{simon:deadends}. While dead-ends in the all-outcome relaxation of FOND problems \cite{ff-replan} are dead-ends of the FOND problem, the reverse
is not true.

\section{Background}
\label{sec:background}

We review classical, generalized, and  FOND planning.

\subsection{Classical planning}

A classical planning problem is a pair $P = \langle D, I\rangle$, where $D$
is  a first-order \emph{domain} and $I$ contains information about
a domain \emph{instance}~\cite{geffner:book,ghallab:book,haslum:pddl}.
The domain $D$ is a set  of action schemas involving a number of domain predicates. 
The action schemas have  preconditions and positive  effects expressed by 
atoms $p(x_1, \ldots, x_k)$ and the negative (delete) effects are negations of such atoms,
where $p$ is a predicate symbol of arity $k$, and each term $x_i$ is a schema argument.

The instance information is a tuple  $I = \langle O, s_0, G \rangle$
where $O$ is a set of objects (constants); $s_0$ is the initial state, and $G$ is the goal.
The ground atoms $p(o_1, \ldots, o_k)$ in the problem instance $P=\tup{D,I}$ are the atoms $p(x_1, \ldots, x_k)$
that result from replacing the terms $x_i$ by objects $o_i \in O$, and the ground actions result from grounding  the action schemas
in a similar way. The states $s$ are sets of ground atoms; those which are true in the state.
The initial state $s_0$ is a set of ground atoms, while $G$ is a set of ground goal atoms.

A classical planning problem $P = \langle D, I \rangle$
defines a state model  $M = \tup{S,s_0,S_G,Act,A,f}$ where $S$ is the set of states, 
$s_0 \in S$ is the initial state, $S_G \subseteq S$ is the set of  goal states,
$Act$ is a set of (ground)  actions, $A(s) \subseteq Act$ is the set of actions applicable
in the state $s$, and $f(a,s)$ for $a \in A(s)$ is a deterministic state transition function.
In the model  $M(P)$ determined by $P$, the states $s \in S$ are collection of ground atoms from $P$,
$s_0$ is given, $S_G$ contains the states that include  $G$,
$Act$ is the set of ground actions, $a \in A(s)$  if the   preconditions of $a$ are true in $s$,
and $s' = f(a,s)$  if  $a \in A(s)$ and $s'$ contains the positive effects of $a$ and the atoms in $s$ except those
deleted by $a$.

It is convenient to consider \emph{non-deterministic policies} for classical planning  problems instead
of (open loop)  plans. A policy $\pi$ for a problem $P$ is a partial function mapping  states $s$ of $P$
into \emph{sets} $\pi(s)$ of actions from $P$, possibly empty. A $\pi$-trajectory in $P$
is a sequence of states $s_0, \ldots, s_n$ that starts in the initial state of $P$
such that $s_{i+1}=f(a_i,s_i)$ if  $a_i \in A(s_i)$ and $a_i \in \pi(s_i)$. 
The trajectory is \emph{cyclic} if it contains the same state infinitely often
and \emph{maximal} if
\begin{enumerate*}
  \item $s_n$ is the first goal state of the sequence,
  \item it is cyclic and does not contain goal states,
  \item there is no action $a_n$ in both $\pi(s_n)$ and $A(s_n)$, or
  \item $\pi(s_n)$ is undefined.
\end{enumerate*}
The policy $\pi$ solves $P$ if the maximal $\pi$-trajectories all  reach
a goal state of $P$. 

\Omit{
A \emph{plan} for $P$ is a sequence of ground actions $a_0, \ldots, a_n$
that is applicable in $s_0$ and results in a goal state; i.e., the sequence
is a plan if there is a sequence of states $s_0, \ldots, s_{n+1}$ such
that $a_i \in A(s)$, $s_{i+1} = f(a_i,s_i)$, and $s_{n+1} \in S_G$.
In the absence of costs, the cost of a plan is its length, and a plan
is optimal if there is no shorter plan. State transitions $(s,a,s')$ will be written as $(s,s')$
when identity of the action that maps $s$ into $s'$ is not relevant. 
}

\subsection{Generalized classical planning}

Departing slightly from previous work, a \emph{general policy} $\pi$ for a class $\Q$ of
classical instances over the same domain  is taken to be a mapping that assigns a (concrete) policy
$\pi_P$ to each problem $P$ in $\Q$. The general policy $\pi$ \emph{solves} $\Q$
if  $\pi_P$ solves $P$ for each $P$ in $\Q$. 

\Omit{
represents a collection of state transitions $(s,s')$ in
each instance $P$ of $\Q$ that are said to be in $\pi$. A $\pi$-trajectory is sequence of states $s_0, \ldots, s_n$
that starts in the initial state of $P$ and whose transitions $(s_i,s_{i+1})$ are all in $\pi$. The trajectory is
maximal if $s_n$ is the first goal state of the sequence,  there is no transition $(s_n,S)$ in $\pi$,
or the trajectory is cyclic and does not contain goal states.
The policy $\pi$ solves $P$ if all maximal $\pi$-trajectories in $P$ reach the goal, and it solves $\Q$ if it solves each
$P$ in $\Q$.
}

A general policy $\pi$ can be  represented in many forms from formulas or rules to value functions.
Following \cite{bonet:ijcai2018,bonet:aaai2019}, we consider general  policies $\pi$ for classes of problems $\Q$
expressed by sets of rules $C \mapsto E$ in terms of a collection $\Phi$ of Boolean  features $p$
and numerical features $n$ that take value in  the non-negative integers. 
The condition $C$  is a set (conjunction) of Boolean feature conditions
and the effect description $E$ is a set (conjunction) of feature value changes.
A Boolean feature condition is of the form $p$, $\neg p$, $n=0$, and $n>0$
for Boolean and numerical features $p$ and $n$ in $\Phi$, and feature value changes
are of the form $p$, $\neg p$, $p?$ for Boolean $p$,   and $\DEC{n}$, $\INC{n}$, and $\UNK{n}$ for numerical $n$.

The general policy $\pi$ for a class of problems $\Q$  defined by a set $R$ of rules $C \mapsto E$
determines for each problem $P$ in $\Q$ the policy $\pi_P$ that maps a reachable state $s$ in $P$
into the set of actions $\pi_P(s)$,  where $a \in \pi_P(s)$ iff $a$ is applicable in $s$, $a \in A(s)$,
and  the successor state $s'=f(a,s)$ is such that the  transition $(s,s')$ \emph{satisfies} a rule in $R$.
The transition $(s,s')$ satisfies a rule $C \mapsto E$ if 
all  feature conditions in $C$ are  true in $s$, and 
the values of the features change from $s$ to $s'$ according to $E$; i.e.,
if $p$ (resp.\ $\neg p$) is in $E$, then $p(s')=1$ (resp.\ $p(s')=0$),
if $\DEC{n}$ (resp.\ $\INC{n}$) is in $E$, $n(s)>n(s')$ (resp.\ $n(s)<n(s'))$,
if $p$ (resp.\ $n$) is not mentioned at all in $E$, $p(s)=p(s')$ (resp.\ $n(s)=n(s')$),
and if $\EQ{n}$ (resp.\ $\GT{n}$) is in $E$, $n(s')=0$ (resp.\ $n(s')>0$).
The  transition $(s,s')$ satisfying rule $R$ is also said to be \emph{compatible with rule $R$} and if $(s, s')$ is  compatible with some rule $R$ of policy $\pi$, it is called \emph{compatible with the policy $\pi$}.

Methods for learning rule-based general policies for classical planning  from small training instances
have been developed \cite{bonet:aaai2019,frances:aaai2021}.
For this, a set of rules involving a set of features of minimum complexity is obtained
by finding a satisfying assignment to a propositional theory $T({\cal S},{\cal F})$
of  minimum cost, where ${\cal S}$ is the collection of state transitions appearing in the
training instances, and ${\cal F}$ is a large pool of features obtained from the domain predicates
in a domain-independent manner using a  description logic grammar \cite{description-logics}.
The complexity of feature $f$ in  ${\cal F}$ is given by the number of grammar rules
needed to generate the unary predicate $p(x)$ associated with $f$. Such unary predicate
gives  rise to the numerical feature $n_p$ whose value in a state $s$ is given by the number
of objects $o$ for which $p(o)$ is true in $s$, and the Boolean feature $b_p$ that is true in $s$
if $n_p$ is positive in $s$. Since problems $P$  in the target class $\Q$ often have different goals,
it is assumed that the  states $s$ in $P$  are extended with a suitable ``copy'' of the goal atoms;
for each goal atom  $p(o_1, \ldots, o_k)$, the states $s$ in $P$  are extended with the
atom $p_G(o_1, \ldots, o_k)$ where $p_G$ is a new predicate \cite{martin:generalized}.

\subsection{FOND Planning}

A FOND model is a tuple $M = \tup{S,s_0,S_G,Act,A,F}$ similar to the one  underlying classical planning
except that the state transition function $F$ is non-deterministic and maps an action $a$ applicable in
a state $s$ into a non-empty set of successor states $s' \in F(a,s)$.  The syntax for FOND problems is
an  extension of the  syntax for classical planning where the actions $a \in A$ are \emph{sets} 
$a=\{b_1, \ldots, b_k\}$ of classical, deterministic actions $b_i$, all  sharing the same preconditions.
The application of $a$ results in the random application of one of the actions $b_i$ 
so that if $a \in A(s)$, $F(a,s) = \{f(b_1,s), \ldots, f(b_k,s)\}$.
A (non-deterministic) policy $\pi$ for a FOND problem $P$ is a partial function  that
maps states into sets of actions of $P$. The $\pi$-trajectories $s_0, \ldots, s_n$  for FOND problems $P$
are defined in the same way as for classical problems except  that for each $a_i \in \pi(s_i)$, the condition $s_{i+1}=f(a_i,s_i)$ is replaced by $s_{i+1} \in F(a_i,s_i)$.
In addition, a notion of \emph{fairness} is needed in FOND planning that can be specified by  considering
$\pi$-trajectories that  include the actions  as $s_0,a_0,s_1,a_1, \ldots, s_n$ where $a_i  \in \pi(s_i)$.
One such  trajectory is deemed \emph{fair} if it is finite, or if it is infinite, and
infinite occurrences of states $s_i$ followed by the same action $a_i$
are  in turn followed by each of the possible successor states $s_{i+1} \in F(a_i,s_i)$ an infinite number of times.
A policy $\pi$ is  a  \emph{strong cyclic solution} or simply a \emph{solution} of  $P$
if  the maximal  $\pi$-trajectories that are \emph{fair} all reach the goal.

\subsection{Dead-ends and deterministic relaxations} 

A state $s$ is reachable in a classical or FOND problem $P$ if there is a trajectory $s_0, \ldots, s_n$ that reaches $s$,
where $s=s_n$ and $s_{i+1}=f(a_i,s_i)$  or $s_{i+1} \in F(a_i,s_i)$ for $i=0, \ldots, n-1$ and  suitable actions $a_i$ in $P$.
For a reachable state  $s$ in $P$, $P[s]$ defines the problem that is like $P$ 
but with initial state $s$. A reachable state $s$ in $P$ is \emph{alive} if $P[s]$ has a solution and a \emph{dead-end} otherwise.
Since a general policy $\pi$ is often aimed at  solving  all solvable instances $\Q$ in  a given domain, 
it is natural to ask for the class $\Q$ to be closed, in the sense that if $P$ is in $\Q$, then $P[s]$ is in $\Q$ if $s$ is not
a dead-end. The set of dead-ends in a FOND problem $P$ is related to the set of dead-end states in the classical problem $P_D$
that results  from $P$ when  each non-deterministic action $a=\{b_1,\ldots,b_m\}$ is replaced by the set of  deterministic actions $b_1, \ldots, b_m$.
The classical problem  $P_D$ is the so-called deterministic relaxation or all-outcome relaxation \cite{ff-replan}
and it  plays an important role in   FOND planners that rely on classical planning algorithms  \cite{prp}.
Clearly, if $s$ is a dead-end state in $P_D$, $s$ will be a  dead-end state in the FOND problem $P$,
but the inverse implication is not true. 

\Omit{
Interestingly, we can prove the following theorem. For a FOND problem $P$, $\Q_P$ denote  the collection of FOND problems that includes
$P$ and $P[s]$ for any state reachable in $s$ that is a reachable state that is not a dead-end. Let $\Q_{P}^D$ denote the collection of classical problems $P_D$
for each FOND problem $P$ in $\Q$. 

\medskip

\hector{Below part: to be worked out; not in Background though:}

\medskip

\noindent \textbf{Theorem:}  There is a  policy $\pi$ that solves the  collection of FOND problems  $\Q_P$ iff there is a general policy $\pi'$
for the collection of classical  problems $\Q_P^D$.

\medskip

\noindent \textbf{Idea:} A) Construct $\pi'$ from from $\pi$, B) Construct $\pi$ from $\pi'$.

\medskip

\noindent A) For each $s$, pick successor $s'$ of $\pi(s)$ in $s$ that is closest to the goal (min-length $\pi$-trajectory from $s$ to goal must go through one such successor). Add this transition to $\pi'$. Need to show then that $\pi'$ solves $\Q_P^D$ if $\pi$ solves $\Q_P$. B) To be continued.

Point is how best to motivate our formulation of \textbf{general FOND policies} in terms of \textbf{general classical policies} and \textbf{safety constraints} (that do not have to be dead-ends necessarily).

}

\section{General Policies for FOND Planning}%
\label{sec:general-fond-policies}

We consider the semantics of general FOND policies and  the language to describe them.

\subsection{Semantical considerations}

The \emph{semantics} of general policies for classes $\Q$ of FOND problems is clear and direct:
a general policy $\pi$ for $\Q$ must determine a policy $\pi_P$ for each problem $P$ in $\Q$,
and $\pi$ solves $\Q$ if each problem $P$ in $\Q$ is solved by $\pi_P$; i.e., if $\pi_P$
is a strong cyclic policy for $P$. The \emph{language} for representing general policies
for classes of FOND problems, however, is a bit more subtle than in the case of classical planning.
Nonetheless, a tight  relation between general policies for FOND problems and general policies for classical problems
can be established that will serve to motivate the language for expressing and then learning
general FOND policies.

Let $\Q$ be a collection of solvable FOND problems $P$ that is \emph{closed} in the following
sense: if $P$ is in $\Q$ and $s$ is an \emph{alive state} reachable in $P$, then $P[s]$ is also in $\Q$.
Let $\Q_D$ stand for the \emph{determinization} of $\Q$; namely, the collection of  classical problems $P_D$
obtained from the deterministic (all-outcome) relaxation of the  FOND problems $P$ in $\Q$. Let us
also say that a general policy $\pi_D$ for the determinization $\Q_D$ of $\Q$ is \emph{safe}
in $P_D$ if for every reachable state $s$ of $P_D$ and every (deterministic) action $b_i \in \pi_D(s)$, there is a (non-deterministic) action $a$ such that $b_i \in a$ and no $s' \in F(a, s)$ is a dead-end.
Finally, $\pi_D$ is \emph{safe} in $\Q_D$ if it is safe in every $P_D \in \Q_D$.
We  can show the following
relation  between the general  policies that solve the class of FOND problems $\Q$
and the general policies that solve the class of classical problems $\Q_D$:\footnote{\ifthenelse{\boolean{techreport}}{Proofs can be found in the appendix.}{Proofs can be found in~\cite{hofmannLearningGeneralizedPolicies2024}.}}

\begin{theoremE}
  \label{thm:gen-policies}
  Let $\Q$ be a collection of solvable FOND problems $P$  that is closed,  and let $\Q_D$ be
  determinization of $\Q$.
  \begin{enumerate}
    \item If $\pi$ is a general policy   that solves the  FOND problems $\Q$,
      a general safe policy $\pi^D$ can be constructed from $\pi$   that solves the class of classical problems $\Q_D$.
    \item If $\pi^D$ is a general safe policy  that solves the classical problems $\Q_D$, a general policy $\pi$
      that solves the  FOND problems $\Q$ can be constructed from $\pi^D$.
  \end{enumerate}
\end{theoremE}
\begin{proofE}
  In the following, let $\pi_P$ be the concrete policy for some $P \in \mathcal{Q}$ obtained from the general policy $\pi$.
  Similarly, let $\pi^D_P$ be the concrete policy for some $P_D \in \mathcal{Q}_D$ obtained from the general policy $\pi^D$.
  \begin{enumerate}
    \item
      We construct $\pi^D$ from $\pi$ as follows.
      Let $P \in \mathcal{Q}$ be any instance of $\mathcal{Q}$ and $P_D$ its determinization.
      Let $s$ be any reachable state in $P$ and $\pi_P(s) = a$ with $a = \{ b_1, \ldots, b_k \}$.
      For each $i$, let $\tau_i = s b_i s_1 \cdots s_{g,i}$ the shortest $\pi_P$-trajectory starting in $s$ with first action $b_i$ and ending in some goal state $s_{g,i}$.
      Then, set $\pi^D_P(s) = b_j$ where $\tau_j$ has minimal length among all $\tau_i$. 

      We need to show that $\pi^D$ is safe and solves $\mathcal{Q}_D$.
      First, suppose $\pi^D$ is not safe and hence for some $P_D \in \mathcal{Q}_D$, there is some $s$ such that $\pi^D_P(s) = b_i$ and for every non-deterministic action $a$ with $b_i \in a$, there is a $s' \in F(a, s)$ such that $s'$ is a dead-end in the FOND problem $P$.
      But then $\pi_P$ may also reach $s'$ and so $\pi_P$ does not solve $P$ and so $\pi$ does not solve $\mathcal{Q}$, contradicting the assumption.

      Now, suppose there is an instance $P_D$ with initial state $s_0$ not solved by $\pi^D_P$.
      Let $\tau$ be a maximal $\pi^D_P$-trajectory starting in $s_0$.
      Clearly, as $\pi^D$ is safe, $\tau$ cannot end in a dead-end state and so must be infinite.
      But then, as $\pi^D_P$ selects the action with minimal distance to a goal, every $\pi_P$-trajectory must be infinite, and so $\pi_P$ does not solve $P$, contradicting the assumption.

      Hence, $\pi^D$ is safe and solves $\mathcal{Q}_D$.
    \item
      We construct $\pi$ from $\pi^D$ as follows.
      Let $P \in \mathcal{Q}$ be any instance of $\mathcal{Q}$ and $P_D$ its determinization.
      Let $s$ be any reachable non-goal state in $P$ and $\pi^D_P(s) = b$.
      As $\pi_D^P$ is safe, there must be some non-deterministic action $a \in A(s)$ with $b \in a$ and such that no $s' \in F(a, s)$ is a dead-end.
      For this action $a$, set $\pi^P(s) = a$.

      We show by contradiction that $\pi$ solves the FOND problems $\mathcal{Q}$.
      Suppose there is an instance $P \in \mathcal{Q}$ with initial state $s_0$ such that $\pi^P$ does not solve $P$ and so there is a $\pi^P$-trajectory $\tau = s_0 \cdots s_n$ ending in some dead-end state $s_n$.
      Let $s_i$ be the last alive state of $\tau$.
      As $\mathcal{Q}$ is closed, $P[s_i] \in \mathcal{Q}$ and so $\pi_P^D(s_i)$ is defined, say $\pi^D_P(s_i) = b$.
      As $\pi^D$ is safe, there must be a non-deterministic action $a$ with $b \in a$ such that no $s' \in F(a, s_i)$ is a dead-end.
      As $\pi_P$ is constructed from $\pi_P^D$, $s_{i+1}$ cannot be a dead-end, contradicting the assumption that $s_i$ is the last alive state of $\tau$.
      Hence, no $\pi^P$-trajectory may end in a dead-end state and so $\pi$ solves $\mathcal{Q}$.
      \qedhere
  \end{enumerate}
\end{proofE}

This result expresses   a basic intuition and the conditions that make the intuition valid; namely,
that  the uncertainty in the action effects of FOND problems can be ``pushed'' as uncertainty in the set of possible initial states,
resulting  in a collection of classical problems, and hence, a generalized classical planning problem.
This suggests that one way to get a general policy for a class $\Q$ of FOND problems is by finding a general  policy for the classical problems
in the determinization $\Q_D$. The theorem qualifies this intuition by requiring that the policy that solves $\Q_D$ must be safe
and not visit a dead-end state of $P$, because a state may be a dead-end  in $P$ but not in
its determinization $P_D$. The intuition that FOND planning can leverage classical planning in this way is present in a slightly
different form in one of the most powerful FOND planners \cite{prp}. The correspondence between FOND and classical planning
can be captured more explicitly in the generalized planning setting as a FOND problem does not map into a single classical planning
problem but into   a collection of them.
\Omit{
The use of a collection of FOND problems $\Q$ in Theorem~\ref{thm:gen-policies}
instead of a single one is a convenience that  will useful for defining a language for describing  general  FOND policies. 
}

\subsection{Expressing general FOND policies}

The correspondence captured by Theorem~\ref{thm:gen-policies} implies that
general policies $\pi$ for a class of FOND problems $\Q$ can be obtained from the
general policies $\pi'$  for $\Q_D$ that are safe, i.e., those policies that avoid dead-ends in the ``original'' FOND problem $P$.

This observation suggests that a suitable language for defining general FOND policies
can be obtained by combining the rule language for describing general policies
for classical domains with \emph{constraints} that ensure that the general
policies that solve the classical problems $\Q_D$ are safe and do not visit
dead-end states of the FOND problem:

\begin{definition}
  The language for representing a \emph{general  policy} over a class $\Q$ of FOND problems
  is made up of a set $R$ of rules $C \mapsto E$ like for general classical policies,
  and a set of constraints $B$, each one being an (implicit) conjunction of Boolean
  feature conditions like $C$. 
\end{definition}

Both the rules $R$  and the constraints $B$ are defined over a set $\Phi$ of Boolean
  and numerical features that are well defined over the reachable states of the problems $P \in \Q$.
The \emph{general FOND policy} defined by a pair of rules $R$ and constraints $B$ is as follows:

\begin{definition}
  A set of rules $R$ and constraints $B$ define a general FOND policy $\pi=\pi_{R,B}$ over $\Q$
  such that in a problem $P \in \Q$,  the concrete policy  $\pi_P$ is such that 
  $a \in \pi_P(s)$ iff 
  \begin{itemize}
  \item there is a state $s' \in F(a,s)$ such that the transition $(s,s')$ satisfies a rule $C \mapsto E$ in $R$, {and}
   \item there is \emph{no} state $s' \in F(a,s)$ such that $s'$ satisfies a constraint in $B$.
  \end{itemize}
  \label{def:pi-fond}
\end{definition}

\noindent Let us say  that a set of constraints $B$ is \emph{sound} relative to a class of FOND problems $\Q$
if every reachable dead-end state $s$ in a problem $P$ in $\Q$ satisfies a constraint in $B$.
Furthermore, a general classical policy $\pi$ is \emph{$B$-safe} if for every reachable state $s$ and every (deterministic) action $b_i \in \pi(s)$, there is a (non-deterministic) action $a$ such that $b_i \in a$ and no $s' \in F(a, s)$ satisfies a constraint in $B$.
The  basic idea of the method for learning general FOND policies that we will pursue
can then be expressed as follows:
\begin{theoremE}
  Let  $\Q$ be a class of FOND problems, $\Q_D$ its determinization, and $B$ a sound set of constraints
  relative to $\Q$. If the rules $R$ encode a general classical policy that solves $\Q_D$ which is $B$-safe,
  then the general FOND policy $\pi_{R,B}$ that follows from  Definition~\ref{def:pi-fond} solves $\Q$.
  \label{thm:key-idea}
\end{theoremE}
\begin{proofE}
  By contraposition.
  Suppose $\pi^D$ is a general classical policy encoded by rules $R$ that is $B$-safe and let $B$ be sound relative to $\mathcal{Q}$. 
  Assume $\pi = \pi_{R,B}$ does not solve $\Q$ and so there is a $P \in \Q$ such that the corresponding concrete policy $\pi_P$ does not solve $P$, i.e., there is a maximal $\pi_P$-trajectory $\tau$ that is fair but does not reach the goal.

  Suppose $\tau$ is finite and so it ends in a non-goal state $s$ such that $\pi_P(s) = \emptyset$ and so for every action $a \in A(s)$, if there is $s' \in F(a, s)$ satisfying some rule in $R$, then there must be $s'' \in F(a, s)$ that satisfies some constraint in $B$.
  As $\pi_D$ is $B$-safe and follows the same rules $R$, this also means that $\pi_P^D(s) = \emptyset$ and so $\pi_D$ does not solve $P_D$.

  Now, suppose $\tau$ is infinite and so contains a cycle $c = s_0 \cdots s_n s_0$.
  Wlog, assume $c$ is the largest cycle in $\tau$.
  First, note that all states in $c$ must be alive: Otherwise, suppose $s_j$ is the last alive state in $\tau$.
  As $\pi^D$ is $B$-safe, $B$ is sound relative to $\Q$, and none of the states $s' \in F(\pi_P(s_j), s_j)$ satisfy a constraint in $B$, no such $s'$ can be a dead-end.
  Hence, as $\Q$ and $\Q_D$ is closed, it follows for each $s_i$ of $c$, as $s_i$ is alive, that $P[s_i] \in \Q_D$.
  Furthermore, as $\tau$ is fair, for every $s_i$ of $c$ and every $a \in \pi_P(s_i)$, all successors $F(a, s_i)$ are in $c$, as otherwise $\tau$ would eventually leave the cycle and thus be finite.
  Now, note that every $\pi_P^D$-trajectory is also a $\pi_P$-trajectory and so every $\pi^D_P$-trajectory visiting some state of $c$ is infinite.
  Hence, for each $s_i$ of $c$, $\pi^D$ does not solve $P_D[s_i]$ and so $\pi^D$ does not solve $\Q_D$.


  Therefore, in either case, $\pi^D$ does not solve $\Q_D$. 
\end{proofE}

\section{Learning General FOND Policies}%
\label{sec:learning}

Following Theorem~\ref{thm:key-idea}, we will  learn general policies $\pi_{R,B}$ that solve
classes of FOND problems $\Q$ as follows: we sample a subclass of small FOND problems $\Q'$ from $\Q$ and
learn rules $R$ and constraints $B$ such that 
the general policy $\pi_R$ solves  the classical problems in  $\Q'_D$
and  is $B$-safe for a sound set of constraints  $B$.
With \autoref{def:pi-fond}, we then obtain a general FOND policy $\pi_{R,B}$ that
solves the FOND  problems in $\Q'$ (but not necessarily all FOND problems in the target class $\Q$).
By looking for the simplest such policies in terms of the cost of the features involved,
we will see that general policies that solve $\Q$ can be obtained.

\subsection{Min-Cost SAT Formulation}

Following \cite{frances:aaai2021,bonet:aaai2019}, the problem
of learning a general policy for a class of classical problems $\Q'_D$ is cast as a combinatorial optimization
problem, and more specifically as min-cost SAT problem over a propositional theory $T = T(\mathcal{S}, \mathcal{F})$
where $\S$ is the set of (possible) state transitions $(s,s')$ over the instances $P_i$ in $\Q$ with states $S_i$, and
$\F$ is the pool of features constructed from predicates in the common domain of these instances. 
The policy rules $R$ are then extracted from the transitions $(s,s')$ that are labeled
as ``good'' in the min-cost satisfying assignment of $T$ by looking at how
the selected features change across the transitions. The constraints $B$ will be 
extracted from $T$ by enforcing a separation between the states that
are dead-ends in $\Q'$ from those that are not. The states appearing in $\mathcal{S}$
are pre-partitioned into alive, dead-end, and goal states, as explained below in Section~\ref{sec:dead-end}.

The cost of an assignment is given by adding the costs of the features selected from the pool $\F$.
Every  feature  $f \in \mathcal{F}$ has  a weight $w(f)$ defined by the number of
grammar rules needed to  derive the unary  predicate $p(x)$ that defines $f$.
The numerical feature $n_p$ expresses  the number of grounded $p(o)$ atoms in a state $s$  (i.e.,
the number of objects that satisfy $p$ in $s$), while
the Boolean feature $b_p$ is true if $n_p$ is positive. 

\Omit{
  The purpose of the propositional theory is to
\begin{enumerate*}
  \item select features such that \emph{dead-ends} can be identified,
  \item from the transitions of $\mathcal{S}$, select \emph{good} transitions that lead to a goal state.
\end{enumerate*}
We assume that $\mathcal{S} = S_1 \cup \ldots \cup S_k$ is the joint spate space of the training instances $P_1, \ldots, P_k$ that is partitioned into alive, dead, and goal states $\mathcal{S} = \alive \dot\cup \dead \dot\cup \goal$ and we denote the solvable states as $\solvable = \alive \cup \goal$.
}

\smallskip

\noindent The \textbf{propositional variables} in $T(\mathcal{S}, \mathcal{F})$ are the following:
\begin{itemize}
  \item $\mi{Good}(s, s')$ is true if the transition $(s, s')$ is good,
  \item $\mi{Select}(f)$ is true if the feature $f$ is selected,
  \item $V(s, d)$ is true if the distance of $s$ to a goal is at most $d$, where $0 \leq d \leq |S_i|$ for $s \in S_i$.
\end{itemize}

\smallskip

\noindent The \textbf{formulas} in  $T(\mathcal{S}, \mathcal{F})$ are  in  turn:

\begin{enumerate}
  \item
    \label{sat:select-good}
    For every alive state $s$: 
    \[
      \bigvee_{a \in \mi{Safe}(s)} \bigvee_{s' \in F(a, s)} \mi{Good}(s, s')
    \]
    where $a \in \mi{Safe}(s)$ if no $s' \in F(a, s)$ is a dead-end.
  \item
    \label{sat:goal-state-zero-distance}
    For every goal state $s$: $V(s, 0)$
  \item
    \label{sat:distance}
    For every alive state $s$: $\exone_{d \in \mathbb{N}}: \{ V(s, d) \}$
  \item
    \label{sat:decrease-distance}
    For every transition $(s, s')$:
    \begin{multline*}
      \mi{Good}(s, s') \wedge V(s, d) \rightarrow
      \\
      \bigwedge_{\substack{a \in A(s): \\ s' \in F(a, s)}} \bigvee_{s'' \in F(a, s)} V(s'', d'') \rightarrow  d'' < d
    \end{multline*}
  \item
    \label{sat:avoid-deadends}
    For every alive state $s$ and dead state $s'$: $\neg \mi{Good}(s, s')$
  \item
    \label{sat:distinguish-goal-states}
    For every goal state $s$ and non-goal state $s'$:
    \[
      \bigvee_{f: \llbracket f(s) \rrbracket \neq \llbracket f(s') \rrbracket} \mi{Select}(f)
    \]
  \item
    \label{sat:distinguish-dead-ends}
    For every alive state $s$ and dead state $s'$:
    \[
      \bigvee_{f: \llbracket f(s) \rrbracket \neq \llbracket f(s') \rrbracket} \mi{Select}(f)
    \]
  \item 
    \label{sat:distinguish-transitions}
    For all transitions $(s_1, s_1')$ and $(s_2, s_2')$:
    \begin{multline*}
      \mi{Good}(s_1, s_1') \wedge \neg \mi{Good}(s_2, s_2') \rightarrow
      \\
      D(s_1, s_2) \vee D2(s_1, s_1', s_2, s_2')
    \end{multline*}
    where
    \[
      D(s_1, s_2) = \bigvee_{f:\llbracket f(s_1)\rrbracket \neq \llbracket f(s_2) \rrbracket} \mi{Select}(f)
    \]
    and
    \[
      D2(s_1, s_1', s_2, s_2') = \bigvee_{f:\Delta_f(s_1, s_1') \neq \Delta_f(s_2, s_2')} \mi{Select}(f)
    \]
\end{enumerate}

\noindent The expressions $\llbracket f(s) \rrbracket$ and $\Delta_f(s,s')$ stand for the value of feature $f$ in $s$,
and the way in which the value of $f$ changes in the transition from $s$ to $s'$
(up, down, and  same value, for both Boolean and numerical features). 
The formulas express the following.  For every alive state, there must be a good transition such that the corresponding FOND action is \emph{safe}, i.e., none of the outcomes lead to a dead-end (\ref{sat:select-good}) and such that one good transition leads towards a goal (\ref{sat:goal-state-zero-distance}, \ref{sat:distance}, \ref{sat:decrease-distance}).
A transition leading to a dead-end may never be good (\ref{sat:avoid-deadends}).
Furthermore, the selected features must be able to distinguish goal from non-goal states (\ref{sat:distinguish-goal-states}), alive states from dead-ends (\ref{sat:distinguish-dead-ends}) and good from non-good transitions (\ref{sat:distinguish-transitions}).

The satisfying assignments of $T(\mathcal{S}, \mathcal{F})$  yield the rules $R$ and the constraints $B$ such that
$B$ is sound relative to the sampled class $\Q'$ of FOND problems, and the classical policy $\pi_R$ given by the rules $R$
constitute a general policy for the classical problems $\Q'_D$ that is $B$-safe. From  Theorem~\ref{thm:key-idea},
the resulting $\pi_{R,B}$ FOND policy that follows from Definition~\ref{def:pi-fond} solves the collection of FOND problems $\Q'$.

\begin{theoremE}
  The theory $T(\mathcal{S}, \mathcal{F})$ is \emph{satisfiable} iff there is a general FOND policy $\pi_{R,B}$ over the  features in the pool $\mathcal{F}$
  that solves the set of sampled FOND problems $\Q'$, such that the selected features  distinguish  dead, alive, and goal states.
\end{theoremE}
\begin{proofE}
  \textbf{$\Rightarrow$:}
  Let $\sigma$ be a satisfying assignment for $T = T(\mathcal{S}, \mathcal{F})$.
  We first construct the feature set $\Phi$ such that $f \in \Phi$ iff $\sigma \models \mi{Select}(f)$.
  The feature set $\Phi$ distinguishes dead, alive, and goal states: By formula \ref{sat:distinguish-dead-ends}, for each pair of alive state $s$ and dead-end $s'$, there must be a feature $f$ such that $\sigma \models \mi{Select}(f)$ and $\llbracket f(s) \rrbracket \neq \llbracket f(s') \rrbracket$.
  Similarly, for each pair of goal state $s$ and non-goal state $s'$, such a distinguishing feature is selected by formula \ref{sat:distinguish-goal-states}.

  For the policy, we use the following construction: Let $\Phi$ be a set of features, $\mathcal{D}$ a set of states, and $\mathcal{T}$ a set of transitions in $\mathcal{S}$, then the policy $\pi_{\mathcal{T},\mathcal{D}}$ is the policy given by the rules $\Phi(s) \mapsto E_1 \vert \cdots \vert E_m$ and constraints $B$ where
  \begin{itemize}
    \item $s$ is a source state in some transition $(s, s')$ in $\mathcal{T}$;
    \item $\Phi(s)$ is the set of Boolean conditions given by the evaluation of $\Phi$ on $s$, i.e., $\Phi(s) = \{ p \mid p(s) = \top \} \cup \{ \neg p \mid p(s) = \bot \} \cup \{ n > 0 \mid n(s) > 0 \} \cup \{ n = 0 \mid n(s) = 0 \}$;
    \item for state $s_i$ with $s_i \models \Phi(s)$ and transition $(s_i, s_i')$ in $\mathcal{T}$, $E_i$ captures the feature changes for $(s_i, s_i')$:
      $E_i = \{ p \mid \Delta_p(s_i, s_i') = {\uparrow} \} \cup \{ \neg p \mid \Delta_p(s_i, s_i') = {\downarrow} \} \cup \{ \inc{n} \mid \Delta_n(s_i, s_i') = {\uparrow} \} \cup \{ \dec{n} \mid \Delta_n(s_i, s_i') = {\downarrow} \}$;
    \item the constraints $B$ are the feature evaluations of the states in $\mathcal{D}$, i.e., $B = \{ \Phi(s) \mid s \in \mathcal{D} \}$.
  \end{itemize}
  The policy $\pi = \pi_{R,B}$ is the policy $\pi_{\mathcal{T},\mathcal{D}}$ where $\mathcal{T} = \{ (s, s') \in \mathcal{S} \mid \sigma \models \mi{Good}(s, s') \}$ and $\mathcal{D}$ is the set of dead-end states in $\mathcal{S}$.

  We show that $\pi$ solves $\mathcal{Q}'$.
  Let $P \in \mathcal{Q}'$ and $\pi_P$ the corresponding concrete policy.
  First, note that for every dead-end state $s$, there is a constraint $B_i \in B$ with $B_i = \Phi(s)$ and so there is no $\pi_P$-trajectory ending in a dead-end state.
  We now show that for every alive state of $P$ there is a maximal $\pi_P$-trajectory reaching a goal.
  Let $s$ be an alive state of $P$.
  By formula \ref{sat:select-good}, there is at least one safe action $a$ with $\sigma \models \mi{Good}(s, s')$ and $s' \in F(a, s)$.
  By formula \ref{sat:distinguish-dead-ends}, $s'$ cannot satisfy any constraint of $B$ and so $\pi_P(s) \neq \emptyset$.
  Furthermore, by formula \ref{sat:distinguish-transitions}, for every $a \in \pi_P(s)$, there is at least one $s'$ such that $\sigma \models \mi{Good}(s, s')$.
  By formulas \ref{sat:goal-state-zero-distance}, \ref{sat:distance}, and \ref{sat:decrease-distance}, each \emph{good} transition corresponds to an action where one of the outcomes reduces the distance to the goal.
  Hence, there must be a finite maximal $\pi_P$-trajectory starting in $s$ ending in a goal state.
  \\
  \textbf{$\Leftarrow$:}
  Let $\pi = \pi_{R,B}$ a general FOND policy over features $\Phi$ that solves $\mathcal{Q'}$ and such that $\Phi$ distinguishes dead, alive, and goal states.
  For each alive state $s$, let $d_\pi(s)$ be the length of the shortest maximal $\pi$-trajectory starting in $s$ and $V_\pi(s) =1 + \max_{a \in \pi(s)} \min_{s' \in F(a, s)} d_\pi(s')$.
  Note that $d_\pi(s)$ and so $V_\pi(s)$ is well-defined because $\pi$ solves $\mathcal{Q'}$ and so each maximal $\pi$-trajectory ends in a goal state and hence its length is finite.
  We construct an assignment $\sigma$ for the variables in $T = T(\mathcal{S}, \mathcal{F})$ that satisfies $T$:
  \begin{itemize}
    \item $\sigma \models \mi{Select}(f)$ iff $f \in \Phi$;
    \item $\sigma \models \mi{Good}(s, s')$ iff the transition $(s, s')$ is compatible with $\pi$;
    \item $\sigma \models V(s, d)$ iff $d = V_\pi(s)$
  \end{itemize}
  We show that $\sigma$ satisfies the formulas in $T$:
  \begin{enumerate}
    \item For every alive state $s$, as $\pi$ solves $\mathcal{Q}'$, there is at least one transition $(s, s')$ compatible with $\pi$ and so $\sigma \models \bigvee_{a \in \mi{Safe}(s)} \bigvee_{s' \in F(a, s)} \mi{Good}(s, s')$.
    \item By definition, the distance of a goal state to a goal is $0$, satisfying $\sigma \models V(s, 0)$ for every goal state $s$.
    \item As $V_\pi(s)$ is well-defined, $\sigma \models \text{ Exactly-1 } V(s, d)$ for each alive state $s$.
    \item As $\pi$ solves $\mathcal{Q}'$, there must be some $\pi$-trajectory starting with a transition $(s, s')$ 
      moving towards the goal, i.e., $V_\pi(s') < V_\pi(s)$, and hence satisfying
      $\mi{Good}(s, s') \wedge V(s, d) \rightarrow \bigwedge_{a \in A(s): s' \in F(a, s) } \bigvee_{s'' \in F(a, s)} V(s'', d'') \rightarrow  d'' < d$.
    \item As $\pi$ solves $\mathcal{Q}'$, there cannot be any transition $(s, s')$ compatible with $\pi$ that from an alive state $s$ to a dead-end $s'$ and so $\sigma \models \neg \mi{Good}(s, s')$.
    \item By assumption, $\Phi$ distinguishes goal from non-goal states and so $\sigma \models \bigvee_{f: \llbracket f(s) \rrbracket \neq \llbracket f(s') \rrbracket} \mi{Select}(f)$.
    \item By assumption, $\Phi$ distinguishes alive from dead-end states and so $\sigma \models \bigvee_{f: \llbracket f(s) \rrbracket \neq \llbracket f(s') \rrbracket} \mi{Select}(f)$.
    \item Let $(s_1, s_1')$ and $(s_2, s_2')$ be two transitions.
      Clearly, if $(s_1, s_1')$ is compatible with a rule $C \mapsto E$ but $(s_2, s_2')$ is incompatible, then
      \begin{enumerate*}
        \item $\Phi(s_1) \neq \Phi(s_2)$, or
        \item $\Delta_f(s_1, s_1') \neq \Delta_f(s_2, s_2')$ for some $f \in \Phi$.
      \end{enumerate*}
      Otherwise, $(s_2, s_2')$ would be compatible with $C \mapsto E$.
      Hence, $\sigma \models \mi{Good}(s_1, s_1') \wedge \neg \mi{Good}(s_2, s_2') \rightarrow D(s_1, s_2) \vee D2(s_1, s_1', s_2, s_2')$.
      \qedhere
  \end{enumerate}
\end{proofE}

Since  we aim to learn a policy that generalizes beyond the sample instances,  the  sum of the weights $w(f)$ of selected features $f$
is minimized to penalize  overfitting. 
Given a satisfying assignment  $T(\mathcal{S}, \mathcal{F})$, the rules $R$ and the constraints $B$
that define the general FOND policy $\pi_{R,B}$ are extracted as follows. 
First, the  features  $\Phi$ are obtained from the true $\mi{Select}(f)$ atoms.
Then, for each true atom $\mi{Good}(s, s')$,  a  rule $C \mapsto E$ is obtained
where $C$ is the Boolean feature valuation true in $s$ (literals $p$, $\neg p$, $n=0$, or $n > 0$),
and $\inc{n} \in E$ if $\Delta_n(s, s') = {\uparrow}$, $\dec{n} \in E$ if $\Delta_n(s, s') = {\downarrow}$, $p \in E$ if $\Delta_p(s, s') = {\uparrow}$, and $\neg p \in E$ if $\Delta_p(s, s') = {\downarrow}$. Duplicate rules are pruned.
Finally, the  state constraints $B$ are extracted from the Boolean feature evaluations of the dead-end states.

\subsection{Dead-End Detection}%
\label{sec:dead-end}

\begin{algorithm}
  \textbf{Input:} FOND model $M(P) = \la S, s_0, S_G, \mi{Act}, A, F \ra$
  \\
  \textbf{Output:} FOND dead-end set $D \subseteq S$
  \begin{algorithmic}[1]
  \State $D \gets \emptyset$; 
  \Repeat
  \ForAll{$s \in S \setminus D$}
  \ForAll{$a \in A(s)$}
  \If{$F(a, s) \cap D \neq \emptyset$}
  \label{alg:dead-ends:remove-action} \State  Remove $a$ from $A(s)$ 
  \EndIf
  \EndFor
  \EndFor
  \ForAll{$s \in S \setminus D$}
  \If{$\neg \exists \text{path } s \xrightarrow{a_1} \ldots \xrightarrow{a_{k}} s_g .\, a_i  \in A(s_i), s_g  \in S_G$} \label{alg:dead-ends:path-condition}
  \State Add $s$ to $D$
  \EndIf
  \EndFor
  \Until{$D$ does not change}
  \State \Return $D$
\end{algorithmic}
\caption{Dead-End Detection}
\label{alg:dead-ends}
\end{algorithm}

\Omit{
\begin{algorithm}
  \textbf{Input:} FOND Model $M = \la S, s_0, S_G, \mi{Act}, A, F \ra$
  \\
  \textbf{Output:} Dead-ends $D \subseteq S$
  \begin{algorithmic}[1]
  \State $D \gets \emptyset$; $M \gets \emptyset$; $i \gets 0$
  \Repeat
  \State $i \gets i+1$; $D_i \gets \emptyset$
  \ForAll{$s \in S \setminus D$}
  \ForAll{$a \in A(s)$}
  \If{$F(a, s) \cap D \neq \emptyset$}
  \State $M \gets M \cup \{ (s, a) \}$
  \EndIf
  \EndFor
  \EndFor
  \ForAll{$s_1 \in S \setminus D$}
  \If{$\neg \exists \text{path } s_1 \xrightarrow{a_1} \ldots \xrightarrow{a_{k}} s_{g}.\, \forall i.\, (s_i, a_i) \not\in M, s_g \in S_G$} \label{alg:dead-ends:path-condition}
  \State $D_i \gets D_i \cup \{ s \}$
  \EndIf
  \EndFor
  \State $D \gets \bigcup D_i$
  \Until{$D_i \neq \emptyset$}
  \State \Return $D$
\end{algorithmic}
\caption{Algorithm to detect dead-ends.}
\label{alg:dead-ends}
\end{algorithm}
}

To identify the sets $D$ of dead-end states in the sampled FOND problems $P_i$, similar to \cite{danieleStrongCyclicPlanning2000}, we iteratively
exclude  every action $a$ from the set of applicable actions $A(s)$ when a  state $s' \in F(a,s)$ is in $D$,
and place  $s$ in $D$ when  there is no  path from $s$ to the  goal using the  applicable  sets $A(s)$
that result.  The resulting algorithm, shown in \autoref{alg:dead-ends}, is sound and complete:

\begin{theoremE}
  \autoref{alg:dead-ends} is sound and complete, i.e.,  state $s \in D$ iff there is no solution of the FOND problem  $P[s]$.
\end{theoremE}
\begin{proofE}
  ~ \\
  \textbf{$\Rightarrow$:}
  By contraposition.
  Let $s_1$ be a solvable state.
  Assume $\pi$ is a solution for $P[s_1]$ and so there is a maximal $\pi$-trajectory $\tau = (s_1, s_2, \ldots, s_k, s_g)$ from $s$ to a goal $s_g$ with transitions $(s_1, a_1, s_2), (s_2, a_2, a_2), \ldots, (s_k, a_k, s_g)$.
  We show by induction on $k$ that no state of $\tau$ (and hence also $s_1$) is marked as dead-end and none of the $a_i$ is removed from  $A(s_i)$.
  \\
  \textbf{Base case.}
  Clearly, there is $0$-length path from $s_g$ to a goal state, as $s_g$ is a goal state.
  \\
  \textbf{Induction step.}
  Assume $s_{i}, s_{i+1}, \ldots, s_g$ are not marked as dead and for any $j > i$, $a_j \in A(s_j)$.
  As $a_j \in A(s_j)$ for each $j > i$, there is path satisfying the condition in line~\ref{alg:dead-ends:path-condition}, and so $s_i$ is not added to $D$.
  Furthermore, $F(a_i, s_i) \cap D = \emptyset$ (otherwise $\tau$ would not be a $\pi$-trajectory) and so $a_i$ is not removed from $A(s_i)$.
  \\
  \textbf{$\Leftarrow$:}
  By contraposition.
  Let $s$ be a state not marked as dead, $D$ the states marked as dead, and $M$ be the set of unsafe actions, i.e., $(s_i, a_i) \in M$ if $a_i$ is removed from $A(s_i)$ by the algorithm in line \ref{alg:dead-ends:remove-action}.
  We construct a policy $\pi$ that solves $P[s]$ as follows:
  For each $s_1 \in S \setminus D$, if $s_1 \in S_G$, set $\pi(s_1) = \emptyset$.
  Otherwise, as $s_1$ is not in $D$, there is a path $p = s_1 \xrightarrow{a_1} s_2 \xrightarrow{a_2} \ldots \xrightarrow{a_k} s_g$ from $s_1$ to a goal state $s_g$ such that $\forall i.\, (s_i, a_i) \not\in M$.
  Wlog assume that $p$ is the shortest of those paths and thus cycle-free, i.e., no state is repeated in the path.
  Set $\pi(s_1) = a_1$.

  We show that every fair maximal $\pi$-trajectory ends in a goal state.
  Assume there is a $\pi$-trajectory $\tau = (s, s_1, \ldots, s_D)$ that ends in a non-goal state $s_D$.
  By definition of $\pi$, $s_D \in D$, as otherwise $\tau$ would not end in $s_D$.
  As $s \not\in D$ and $s_D \in D$, there is a $k$ such that $s_k \not\in D$ but $s_{k+1} \in D$.
  Hence $F(\pi(s_k), s_k) \cap D \neq \emptyset$.
  But then, $(s_k, \pi(s_k)) \in M$, in contradiction to the construction of $\pi$.
  Now, assume there is a fair infinite $\pi$-trajectory $\tau$ and so there is a cycle $c$ in $\tau$.
  Wlog, assume $c$ is the longest cycle in $\tau$.
  Clearly, there is a state $s$ in $c$ with $\pi(s) = a$ for some $a$ such that there is a path $s \xrightarrow{a} s' \rightarrow \cdots \rightarrow s_g$ ending in a goal state $s_g$ and such such that $s'$ is not in $c$.
  As $s'$ is not in $c$, it can occur in $\tau$ only finitely many times.
  On the other hand, $\tau$ visits each state of $c$ infinitely often and so $\tau$ cannot be fair, a contradiction.
  Therefore, every fair maximal $\pi$-trajectory ends in a state $s' \in S \setminus D$.
\end{proofE}

\section{Evaluation}%
\label{sec:evaluation}

We evaluate  the approach on a number of FOND benchmarks, and analyze some of the learned general policies.\footnote{The source code, benchmark domains, and results are available at \url{http://doi.org/10.5281/zenodo.11171181}.}

\subsection{Experimental Results}%
\label{sub:results}

\begin{table*}
  \centering
  \includegraphics{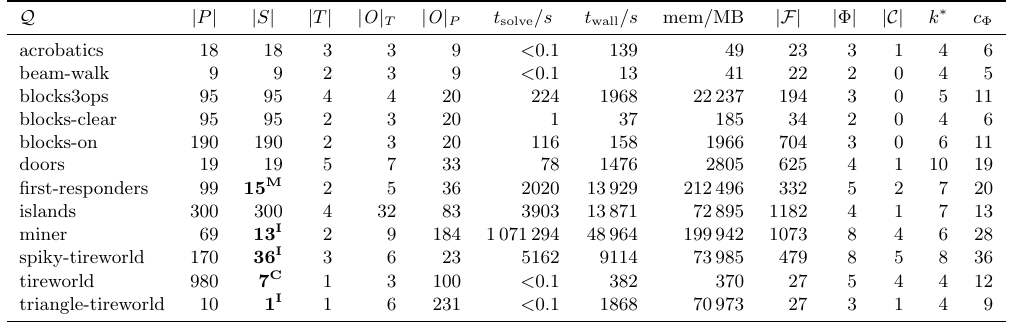}
  \caption{
    Evaluation results, where $|P|$ is the total number of problems, $|T|$ is the number of  problems used in training,
    and $|S|$ is the  number of solved problems, that includes training and testing. $|O|_T$ is the maximum number of objects in all training instances, $|O|_P$ is the maximum number of objects in all instances, $t_\text{solve}$ is the solver's CPU time needed for finding the best policy, $t_\text{wall}$ is the total wall time, $\text{mem}$ is the maximum memory consumption, $|\mathcal{F}|$ is the size of the feature pool, $|\Phi|$ is the number of selected features, $|\mathcal{C}|$ is the number of constraints, $k^*$ is the maximum cost of the selected features, and $c_\Phi$ is the total cost of all selected features.
    When the incremental learning approach does not deliver FOND policies that generalize to all problems in the distribution, the reason for the  failure is indicated:  \textbf{I} indicates that the number of facts exceeded the \texttt{clingo} limits, \textbf{C} indicates that no solution was found with max complexity $15$, and \textbf{M} indicates that the solver ran out of memory.
   }
  \label{tbl:results-state}
\end{table*}


We modeled and solved the min-cost SAT problem represented by the  theory $T(\mathcal{S}, \mathcal{F})$
as an \emph{Answer Set Program} (ASP)~\cite{lifschitzAnswerSetsLanguage2016} in  \texttt{clingo}~\cite{gebserPotasscoPotsdamAnswer2011}.
We use the library \texttt{pddl}\footnote{\url{https://github.com/AI-Planning/pddl}} for PDDL parsing and \texttt{DLPlan} \cite{drexlerDLPlan2022} for feature generation in the same way as \cite{drexler:icaps2022,frances:aaai2021}.
As optimizations, instead of using the ranking $V(s, d)$, we incrementally label all states where all selected transitions lead to the goal as \emph{safe} and require that all \emph{alive} states are also \emph{safe}.
Additionally, we do not try to distinguish all dead states from alive states and instead only compare alive states to \emph{critical} states, which are those states that are dead-ends but have an incoming transition from an alive state.
Finally, we pre-process the state space $S$ by pruning all dead states that are not critical.

The FOND domains considered were  taken from the FOND-SAT distribution  \cite{fondsat}, leaving out domains with unsupported features.
All instances are either randomly generated or taken from the original benchmarks.
In \emph{acrobatics}, \emph{beam-walk}, and \emph{doors}, we augmented the existing problem set with smaller instances.
The problems in the \emph{blocks} variants are generated by scaling from small problems with only three blocks up to 20 blocks.
In \emph{blocks3ops}, the goal is to build a tower of blocks using a three-operator encoding (without a gripper).
The domains \emph{blocks-clear} and \emph{blocks-on} use a four-operator encoding (including the gripper) and the goal is to clear a single block and stack a single pair of blocks.
In \emph{islands}, we created five variations of each problem  from the original problem set.
\emph{Miner} and \emph{triangle-tireworld} use the original problem set, while the instances for \emph{spiky-tireworld} and \emph{tireworld} are randomly generated.
For all domains, the largest generated instances are of similar size or larger than the largest instances in the original benchmarks.

All experiments were run on Intel Xeon Platinum 8352M CPUs with 32 threads, a memory limit of 220 GB, and a maximal feature complexity $c_\text{max} = 15$.
The results are shown in \autoref{tbl:results-state}. The suite of problems $P$ in each domain is ordered by size, with the smallest problems used for training
and the largest problems for testing.
More precisely, starting with a singleton training set consisting of the smallest instance of $P$, the solver learns a new policy and iteratively tests whether the policy solves the next problem.
If this validation fails, the failed instance is added to the training set and the process repeats.
Since the instances in these domains become quite large
and the min-cost SAT solver does not scale up to large instances, if the policies learned from the smallest  instances do not generalize, the approach fails,
as shown by the rows in the table with coverage numbers $|S|$ in bold; namely, 5 of the 12 domains.
In  7 of the 12 domains, on the other hand, the learning method delivers general FOND policies, some of which will be shown to be correct in the next section.

\subsection{Correctness}%
\label{sub:correctness}

For proving the correctness of  learned general FOND policies, we adapt a
method from \cite{frances:aaai2021,seippCorrelationComplexityClassical2016} based on \emph{complete} and \emph{descendent} policies:
\begin{definition}
  A FOND policy $\pi$ is
  \begin{enumerate}
    \item \emph{dead-end-free} if no $\pi$-trajectory visits a dead-end state,
    \item \emph{complete for an instance $P$} if for every alive state $s$, we have $\pi(s) \cap A(s) \neq \emptyset$,
    \item \emph{descending over $P$} if there is some function $\gamma$ that maps states of $P$ to a totally ordered set $\mathcal{U}$ such that for every alive state $s$ and action $a \in \pi(s) \cap A(s)$, we have
  $\gamma(s') < \gamma(s)$ for some $s' \in F(a, s)$.
  \end{enumerate}
\end{definition}

Typically, one can show that a FOND policy $\pi$ is descending by providing a fixed tuple $\la f_1, \ldots, f_n \ra$ of state features.
If for every $\pi$-compatible transition $(s, s')$, we have $\la f_1(s'), \ldots, f_n(s') \ra < \la f_1(s), \ldots, f_n(s) \ra$ with lexicographic order $<$, then $\pi$ is descending.
It can be shown that such a policy indeed solves $P$:
\begin{theoremE}\label{thm:descending-policies}
  If $\pi$ is  a policy that is dead-end-free, complete and descending for an instance $P$, then $\pi$ solves $P$.
\end{theoremE}
\begin{proofE}
  Let a \emph{descending trajectory} be a (finite or infinite) trajectory $\tau = (s_1, s_2, \ldots)$ such that $\gamma(s_{i+1}) < \gamma(s_i)$ for every $i$.
  Clearly, as $\pi$ is descending, for every alive state $s$, there is a maximal $\pi$-trajectory starting in $s$ that is descending.
  Now, suppose there is an alive state $s_1$ such that some descending maximal $\pi$-trajectory $\tau$ starting in $s_1$ does not end in a goal state.
  We have two cases:
  \begin{enumerate}
    \item The trajectory $\tau = (s_1, \ldots, s_k)$ is finite and $s_k$ is not a goal state.
      As $\pi$ is dead-end-free, it follows that $s_k$ is alive and hence, as $\pi$ is complete, there must be some $a \in \pi(s_k) \cap A(s_k)$.
      But then $\tau$ is not maximal.
    \item The trajectory $\tau = (s_1, s_2, \ldots)$ is infinite.
      As the state space of $P$ is finite, the set $\{ \gamma(s_i) \mid s_i \in \tau \}$ has a minimal element.
      Hence, there is an $i$ such that $\gamma(s_{i+1}) \not< \gamma(s_i)$.
      But then $\tau$ is not descending.
  \end{enumerate}
  Hence, for every alive state $s$, there is a descending maximal trajectory starting in $s$ and each such trajectory ends in a goal state.
  Therefore, $\pi$ solves $P$.
\end{proofE}

\subsubsection{Acrobatics}%
\label{ssub:acrobatics-correctness}

An acrobat needs to reach the end of a beam consisting of $n$ segments.
The only ladder to climb up the beam is at its beginning.
The acrobat may walk left or right on the beam and on the ground, climb  up or down if there is a ladder, and jump on the beam.
When walking on the beam, the acrobat may fall down.
The acrobat may skip a segment by jumping over it, but she may fall down and break her leg while doing so.
Once the leg is broken, she may no longer move.

The learned policy $\pi_\text{acro}$ uses three features:
\begin{enumerate*}
  \item the distance $d \equiv \dist(\mi{position}, \mi{next\text{-}fwd}, \mi{position}_G)$ between the current position and the goal position,
  \item a Boolean feature $U \equiv |up|$ which is true if the agent is currently on the beam,
  \item a Boolean feature $B \equiv |\mi{broken\text{-}leg}|$ which is true if the agent's leg is broken.
\end{enumerate*}
The learned policy $\pi_\text{acro} = \pi_{R,B}$ consists of the following rules $R$:\footnote{The notation $C \mapsto E_1 \mid E_2$ abbreviates the two rules $C \mapsto E_1$ and $C \mapsto E_2$ with the same condition $C$.}
\begin{alignat*}{3}
  &r_1:\quad &\{ U, d > 0, \neg B \} &\mapsto \{ \dec{d} \}
  \\
  &r_2:\quad &\{ \neg B, \neg U \} &\mapsto \{ U \} \mid \{ \inc{d} \}
  \\
  \intertext{It has a single constraint $B = \{ b_1 \}$:}
  &b_1:\quad &\{ B, \neg U \}
\end{alignat*}
If the acrobat is currently on the beam ($U$), she is not at the goal ($d > 0$), and the leg is not broken ($\neg B$), then she should decrease the distance to the goal.
Otherwise, if she is not on the beam ($\neg U$) and the leg is not broken ($\neg B$), then she should either climb up the ladder or move away from the goal (and therefore closer to the ladder).
For the first rule, she may decide to jump to decrease the distance and thereby break her leg.
The state constraint forbids this by requiring that she may not end up in a state where she has a broken leg and is not on the beam.
\begin{propositionE}
  The general policy $\pi_\text{acro} = \pi_{R,B}$ solves the class  $\mathcal{Q}_\text{acro}$ of solvable FOND \emph{acrobatics} problems.
\end{propositionE}
\begin{proofE}
  Let $P$ be any $\mathcal{Q}_\text{acro}$ instance and $\pi_P$ the concrete policy for $P$ as defined by $\pi_\text{acro}$.
  First, note that the only critical states are those were the acrobat's leg is broken and she is not on the beam, because in order to break the leg, she needs to fall down with a \texttt{jump}.
  As the state constraint is $\{ B, \neg U \}$, there is no $\pi_P$-compatible transition ending in such a state and so $\pi_P$ is dead-end-free.

  We now show that $\pi_P$ is complete for $P$.
  In every alive state $s \in \alive(P)$ of $P$, we have $B(s) = \bot$.
  Also, $d(s) > 0$ or $U(s) = \bot$.
  If $U(s) = \bot$, then the acrobat is either at the beginning, in which case she can climb up the ladder ($U$), or she may walk back and thereby increase the distance to the end ($\inc{d}$).
  Either way, there is an action compatible with $r_2$.
  Otherwise, if $U(s) = \top$ and the acrobat is not at the goal, then she can continue walking on the beam and  decrease the distance to the end ($\dec{d}$) without violating the constraint.
  Hence, if $U(s)$, then there is a transition compatible with $r_1$ and so $\pi_P$ is complete.

  Finally, we show that $\pi_P$ is descending over tuple $\la 1-U, -(1-U) d, d \ra$.
  First, if the acrobat is on the beam, then $(1-U)$ and $-(1-U)d$ always evaluate to $0$ and $r_1$ is the only applicable rule.
  The only compatible transition is walking toward the end of the beam, decreasing the distance $d$.
  Second, if the acrobat is not on the beam, she can either climb up and decrease the value of $1-U$ from $1$ to $0$, or she may move toward the beginning and decrease the value of $-(1-U)d$ while leaving the value of $1-U$ unchanged.
  Hence, by \autoref{thm:descending-policies}, $\pi_P$ solves $P$.
  As every concrete policy $\pi_P$ solves $P$, it follows that $\pi_\text{acro}$ solves $\mathcal{Q}_\text{acro}$.
\end{proofE}

\subsubsection{Doors}%
\label{ssub:doors-correctness}

The player needs to move through a sequence of $n$ rooms, which are connected by doors.
Whenever the player goes to the next room, the incoming and outgoing doors of the room may open or close non-deterministically.
There are separate actions for moving to the next room depending on whether the door is open or closed.
For the last door, if the door is closed, the player needs to use a key, which is located in the first room.
The player may not move back.

\newcommand*{\playerat}{\ensuremath{\mi{player\text{-}at}}}
\newcommand*{\doorin}{\ensuremath{\mi{door\text{-}in}}}
\newcommand*{\doorout}{\ensuremath{\mi{door\text{-}out}}}
\newcommand*{\finallocation}{\ensuremath{\mi{final\text{-}location}}}
\newcommand*{\holdkey}{\ensuremath{\mi{hold\text{-}key}}}

The learned policy $\pi_\text{doors}$ uses four features:
\begin{enumerate*}
  \item a Boolean feature $G \equiv |\playerat \sqcap \finallocation|$, which is true if the player is at the final location, 
  \item a Boolean feature $S \equiv |\neg \exists \doorin. \playerat|$, which is true if the player is at the start location (which does not have any incoming door),
  \item a Boolean feature $K \equiv | \holdkey |$ which is true if the player is holding the key,
  \item a Boolean feature $F \equiv |\mi{open} \sqcap (\exists \doorout. \playerat) \sqcap \exists \doorin. \finallocation|$, which is true if the player is in the second-last room and the door to the final room is open.
\end{enumerate*}

The policy $\pi_\text{doors} = \pi_{R,B}$ uses the following rules $R$:
\begin{alignat*}{3}
  &r_1: \quad &\{\neg G, S, K, \neg F \} &\mapsto \{ \neg S \}
  \\
  &r_2: \quad &\{\neg G, S, K, F \} &\mapsto \{ G, \neg S, \neg F \}
  \\
  &r_3: \quad &\{\neg G, S, \neg K \} &\mapsto \{ K \} \mid \{ G, \neg S, \neg F \}
  \\
  &r_4: \quad &\{\neg G, \neg S, K, \neg F \} &\mapsto \{ \} \mid \{ F \} \mid \{ G \}
  \\
  &r_5: \quad &\{\neg G, \neg S, \neg K, F \} &\mapsto \{ G, \neg F \}
  \\
  \intertext{It uses one constraint $B = \{ b_1 \}$:}
  &b_1: \quad &\{\neg G, \neg F, \neg S, \neg K \}
\end{alignat*}

The need for feature $F$ may not be immediately obvious, as it is not necessary for a strong-cyclic policy starting in the initial state.
However, it is needed to distinguish dead from alive states, as the state where the player is in the second-last room without a key and the last door is open is also alive: the player may just move through the open door without a key.
Similarly, if $F$ is false and the player is not holding the key, then the state is dead if the player is not at the start location.

We can show that this policy is a solution for $\mathcal{Q}_\text{doors}$:
\begin{propositionE}
  The general policy $\pi_\text{doors} = \pi_{R,B}$ solves the class  $\mathcal{Q}_\text{doors}$ of solvable FOND \emph{doors} problems.
\end{propositionE}
\begin{proofE}
  Let $P$ be any $\mathcal{Q}_\text{doors}$, $\pi_P$ the concrete policy for $P$ as defined by $\pi_\text{doors}$, and $s$ an alive state.
  We first show that $\pi_P$ is dead-end-free.
  It is easy to see that the dead states are exactly those where the player is not holding a key, is not at the start location, and is also not in the second-last room with an open final door, i.e., those states that satisfy $\{ \neg F, \neg S, \neg K \}$, which is the (only) state constraint of $\pi_\text{doors}$ and hence those states will never be visited.

  Next, we show that every maximal $\pi_P$-compatible trajectory starting in $s$ is finite and ends in a goal state.
  As the player may not move back and may also not put down the key, any trajectory and hence also every $\pi_P$-compatible trajectory may not visit the same state more than once.
  Hence, any such trajectory is finite.
  Now, for states with $G(s) = \bot$, notice that for every Boolean combination of $S$, $K$, and $F$ except $\{ \neg F, \neg S, \neg K \}$ and hence for every possible evaluation of an alive state, the policy contains a rule $C_i \mapsto E_i$ such that $C_i$ is satisfied and there is an $E_i$-compatible transition.
  Hence, any maximal trajectory may not end in a state with $G(s) = \bot$.
  As $G(s) = \top$ iff $s$ is a goal state, every maximal $\pi_P$-compatible trajectory starting in an alive state ends in a goal state.
  Hence, for every instance $P$, the corresponding  concrete policy $\pi_P$ solves $P$, and so $\pi_\text{doors}$ is a solution for $\mathcal{Q}_\text{doors}$.
\end{proofE}

\subsubsection{Islands}%
\label{ssub:islands}

In \emph{Islands}, there are two islands connected by a bridge.
The person starts on one island while the goal is on the other island.
They may swim across but with the risk to drown, from which they cannot recover.
Alternatively, they may cross a bridge, but only if there are no monkeys on the bridge.
A monkey can be moved to a drop location.

The learned policy $\pi_\text{islands}$ uses three features:
\begin{enumerate*}
  \item a Boolean feature $A \equiv |\mi{person\text{-}alive}|$;
  \item a numerical feature $d_\mi{drop} \equiv \dist(\mi{bridge\text{-}drop\text{-}location} \sqcap \mi{bridge\text{-}road}[0], \\ \mi{road},\mi{person\text{-}at})$, which is the distance to a location that is both drop location and starting point of the bridge;
  \item a numerical feature $d_g$, which is the distance to the goal: $d_g \equiv \dist(\mi{person\text{-}at}_G,\mi{road},\mi{person\text{-}at})$.
\end{enumerate*}

The policy $\pi_\text{islands} = \pi_{R,B}$ consists of two rules $R = \{ r_1, r_2 \}$:
\begin{alignat*}{3}
  &r_1: \quad &\{ A, d_\mi{drop} = 0, d_g > 0 \} &\mapsto \{ \} \mid \{ \dec{d_g}, \inc{d_\mi{drop}} \}
  \\
  &r_2: \quad &\{ A, d_\mi{drop} > 0, d_g > 0 \} &\mapsto \{ \dec{d_\mi{drop}} \} \mid \{ \dec{d_g }\}
  \intertext{It uses a single constraint $B = \{ b_1 \}$:}
  &b_1: \quad &\{ \neg A, d_\mi{drop} > 0, d_g > 0 \}
\end{alignat*}
The agent first moves to the bridge ($r_2$).
After it has reached the bridge, it directly crosses it if possible ($\{ \dec{d_g}, \inc{d_\mi{drop}} \}$).
Otherwise, it selects an action that does not have any effect on the features ($\{ \}$).
The only action that is compatible with $\{ \}$ is moving a monkey.
As this demonstrates, it is not necessary to encode the monkeys in the policy explicitly.
Finally, the only constraint $b_1$ requires that the person never dies.

\begin{propositionE}
  The general policy $\pi_\text{islands} = \pi_{R,B}$ solves the class  $\mathcal{Q}_\text{islands}$ of solvable FOND \emph{islands} problems.
\end{propositionE}
\begin{proofE}
  Let $P$ be a solver \emph{islands}  instance and $\pi_P$ the corresponding concrete policy according to $\pi_\text{islands}$.
  We first show that $\pi_P$ is dead-end-free.
  The only dead-ends are those where the person is not alive, caused by swimming across the water.
  As swimming never directly leads to the goal (because the player needs to continue moving on the other island) and swimming from the bridge is also not possible, any attempt to swim violates the state constraint $\{ \neg A, d_\mi{drop} > 0, d_g > 0 \}$.
  Hence, $\pi_P$ never reaches a dead-end.

  We now show that $\pi_P$ is complete: Clearly, moving closer towards the bridge and hence decreasing $d_\mi{drop}$ is possible as long as $d_\mi{drop} > 0$.
  Once the person has reached the bridge, it is either free of monkeys, in which case the agent can cross the bridge and thereby decrease $d_g$ while increasing $d_\mi{drop}$.
  Otherwise, if there are monkeys on the bridge, they can always move a monkey, because they are at a drop location, until the bridge is eventually free.
  Note that this assumes that the drop location is actually the same location as the start of the bridge, which is the case for all instances.
  If this were not the case, then the policy would be incomplete.
  After crossing the bridge, they can always move towards the goal until eventually reaching it.

  Now, we show that $\pi_P$ descending over tuple $\la d_g, n_m, d_\mi{drop} \ra$, where $n_m$ is the number of monkeys on the bridge.
  (Note that for tuple $\la d_g, d_\mi{drop} \ra$ and in fact for any tuple over $\Phi$, the policy is not descending, as $r_1$ has an effect that does not change any feature values).
  Clearly, while the person is on the first island away from the bridge, they will decrease distance $d_\mi{drop}$ with every action.
  Once they have reached the bridge, either $d_g$ or $n_m$ will be decreased, until the goal is reached eventually.

  Hence, by \autoref{thm:descending-policies}, $\pi_P$ solves $P$, and so $\pi_\text{islands}$ solves $\mathcal{Q}_\text{islands}$.
\end{proofE}

\section{Variation: Transition Constraints}%
\label{sec:variations}

\begin{table*}
  \centering
  \includegraphics{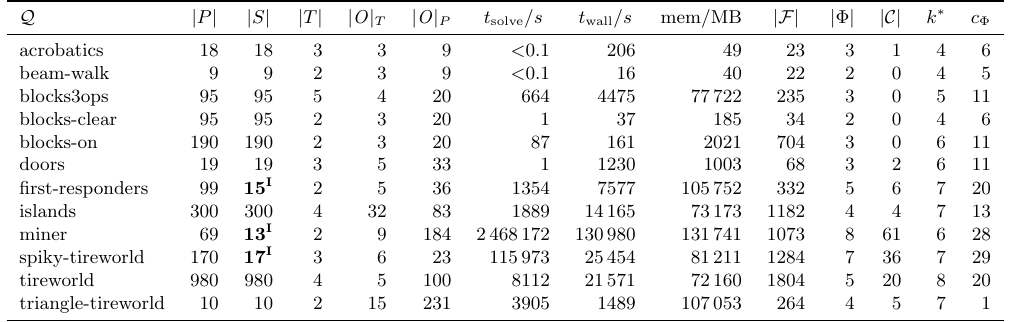}
  \caption{Evaluation results for policy learning with transition constraints, using the same notation as in \autoref{tbl:results-state}.}
  \label{tbl:results-trans}
\end{table*}

The general FOND policies and learning schema presented above is based on \emph{state constraints}, which describe states that must be avoided.
Alternatively, we can also formulate general policies based on \emph{transition constraints}. 
Syntactically, transition constraints are like policy rules and have the form $C \mapsto E$.
However, they describe \emph{bad transitions} and hence the policy  $\pi$ defined by a set of rules and transition constraints
is such that for any $P \in \Q$, $a \in \pi_P(s)$ if the transition $(s',s)$ for some $s' \in F(a,s)$ satisfies a rule,
and no state $s'' \in F(a, s)$ satisfies a transition constraint.
%
%
Formally:
\begin{definition}
  The language for representing a \emph{general  policy with transition constraints} over a class $\Q$ of FOND problems
  is made up of a set $R$ of rules $C \mapsto E$ like for general classical policies,
  and a set of transition constraints $T$ of the same form as rules.
\end{definition}

\begin{definition}
  A set of rules $R$ and transition constraints $T$ define a transition-constrained general FOND policy $\pi$ over $\Q$
  such that  $a \in \pi_P(s)$, where $\pi_P$ is the concrete policy determined by the general policy $\pi$ in  problem $P$ in $\Q$ if 
  \begin{itemize}
  \item there is a state $s' \in F(a,s)$ such that the transition $(s,s')$ satisfies a rule $C \mapsto E$ in $R$, and
  \item there is \emph{no} state $s' \in F(a,s)$ such that the transition $(s, s')$ satisfies a transition constraint $C \mapsto E$ in $T$.
  \end{itemize}
  \label{def:transition-constraints-pi-fond}
\end{definition}

We call a transition $(s, s')$ in a problem $P$ \emph{critical} if $s$ is alive and $s'$ is a dead-end.
Analogously to state constraints, we say  that a set of transition constraints $T$ is \emph{sound} relative to a class of FOND problems $\Q$,
if every critical transition $(s, s')$ in a problem $P$ in $\Q$ satisfies a constraint in $T$,
and that a general  policy $\pi$ for a class of classical or FOND  problems $\Q$ is \emph{T-safe}
if for no instance $P$ in $\Q$, there is a $\pi$-trajectory containing a critical transition.

\begin{theoremE}
  Let  $\Q$ is  a class of FOND problems, let  $\Q_D$ be  its determinization, and let $T$ be a sound set of transition constraints
  relative to $\Q$.  Then  if the rules $R$ encode a general classical policy that solves $\Q_D$ which is $T$-safe,
  the  rules $R$ and  constraints $T$ define a general FOND policy $\pi_{R,T}$ that solves $\Q$.
  \label{thm:transition-constraints-key-idea}
\end{theoremE}
\begin{proofE}
  By contraposition.
  Let $\pi^D$ be a general classical policy encoded by rules $R$. 
  Assume $\pi = \pi_{R,B}$ does not solve $\Q$ and so there is a $P \in \Q$ such that the corresponding concrete policy $\pi_P$ does not solve $P$.

  As $T$ is sound relative to $\Q$ and no transition compatible with $\pi_P$ satisfies a constraint in $T$, $\pi_P$ does not reach any dead-end state.
  Now, suppose there is a $\pi$-reachable non-goal state $s$ such that $\pi_P(s) = \emptyset$ and so for every action $a \in A(s)$, if there is $s' \in F(a, s)$ satisfying some rule in $R$, then there must be $s'' \in F(a, s)$ such that $(s, s'')$ satisfies some constraint in $T$.
  But as $\pi_D$ is $T$-safe and follows the same rules $R$, this also means that $\pi_P^D(s) = \emptyset$ and so $\pi_D$ does not solve $P_D$.

  Therefore, if $\pi_P$ does not solve $P$, there must be a $\pi_P$-reachable state $s$ such that every maximal $\pi_P$-trajectory starting in $s$ is infinite.
  By definition, every trajectory of the classical policy $\pi_P^D$ 
  is also a trajectory of $\pi_P$.
  Hence, every maximal $\pi_P^D$-trajectory starting in $s$ is infinite and so $\pi_P^D$ does not solve $P$.

  Hence, in either case, $\pi^D$ does not solve $\Q_D$. 
\end{proofE}

The experimental results that follow from the use of transition constraints instead of state constraints
for defining and learning general FOND policies are shown in \autoref{tbl:results-trans}.
We can see that in contrast to the state-based variant, the transition-based variant solves all instances of \emph{tireworld} and \emph{triangle-tireworld}.

\section{Conclusion}%
\label{sec:conclusion}

We have extended the formulation for learning general policies for classical planning domains to fully-observable non-deterministic domains.
The new formulation for expressing and learning FOND policies exploits a correspondence between the general policies that solve
a family $\Q$ of FOND problems and the general  safe policies that solve a family $\Q_D$ of classical problems $P_D$ obtained from the all-outcome relaxation (determinization)
of the instances  $P$ in $\Q$, where the safe policies are those that avoid the dead-end states of $P$.
A representation of the collection of dead-end states is learned along with the features and rules. 
The resulting safe policies for the family of classical problems $P_D$ do not just solve the FOND problems in $\Q$ but
potentially many other FOND problems as well, like those that result from random perturbations which do not create new dead states.
This is because the formulation pushes the uncertainty in the action outcomes into uncertainty in the initial states
that are all covered by the general policy that solves $\Q_D$.
The experiments over existing FOND benchmarks show that the approach is sufficiently practical,
resulting in general FOND policies that can be understood and shown to be correct.

\FloatBarrier

\section*{Acknowledgements}

This work has been supported by the Alexander von Humboldt Foundation with funds from
the Federal Ministry for Education and Research.  It has also  received funding from
the European Research Council (ERC), Grant agreement No 885107, the Excellence Strategy
of the Federal Government and the NRW Länder, Germany, and
the Knut and Alice Wallenberg (KAW) Foundation under the WASP program.

{ \small
\bibliographystyle{named}
\bibliography{zotero,control}
}

\ifthenelse{\boolean{techreport}}{

\clearpage
\appendix

\section{Feature Pool}
\label{sec:feature-pool}

\newcommand*{\inter}{\ensuremath{s}\xspace}
The feature pool is constructed iteratively based on a description logic grammar~\cite{dl-handbook} using \texttt{DLPlan}~\cite{drexlerDLPlan2022} similar to \cite{frances:aaai2021}.

\subsection{Description Logic Concepts and Roles}%

In description logic, \emph{concepts} represent unary relations and \emph{roles} represent binary relations.
Higher-arity domain predicates can be represented by concepts and roles as follows: For each $k$-ary domain predicate $p$, we add a primitive concept $p[i]$ for $0 \leq i < k$ that denotes the $k$-th argument of $p$.
Similarly, we add a primitive role $p[i, j]$ for $0 \leq i, j < k$ that denotes the pair $(p_i, p_j)$ of the $i$th and $j$th argument of $p$.

Here, we define the semantics directly in terms of a planning state $s$, where the universe $\Delta^s$ consists of the set of objects occurring in $s$ and the semantics of primitive concepts and roles is defined as follows.
For every $k$-ary state predicate $p$ and $0 \leq i, j < k$:
\begin{itemize}
  \item $(p[i])^\inter = \{ c_i \in \Delta^\inter \mid p(c_0, \ldots, c_i, \ldots, c_{k-1}) \in s \}$,
  \item $(p[i, j])^\inter = \{ (c_i, c_j) \in \Delta^\inter \times \Delta^\inter \mid p(c_0, \ldots, c_i, \ldots, c_j, \ldots, c_{k-1}) \in s \}$.
\end{itemize}

We continue with the compositional roles and concepts.
Let $C, D$ be concepts and $R, S$ roles.
We iteratively construct the following compositional concepts:
\begin{itemize}
  \item the universal concept $\top$ where $\top^\inter = \Delta^\inter$,
  \item the bottom concept $\bot$ where $\bot^\inter = \emptyset$,
  \item intersection $C \sqcap D$ where $(C \sqcap D)^\inter = C^\inter \cap D^\inter$,
  \item union $C \sqcup D$ where $(C \sqcup D)^\inter = C^\inter \cup D^\inter$,
  \item negation $(\neg C)$ where $(\neg C)^\inter = \Delta^\inter \setminus C^\inter$,
  \item difference $(C \setminus D)$ where $(C \setminus D)^\inter = (C^\inter \setminus D^\inter)$,
  \item existential restriction $\exists R.C$ where $(\exists R.C)^\inter = \{ a \mid \exists b:\: (a, b) \in R^\inter \wedge b \in C^\inter \}$,
  \item universal restriction $\forall R.C$ where $(\forall R.C)^\inter = \{ a \mid \forall b:\: (a, b) \in R^\inter \rightarrow b \in C^\inter \}$,
  \item constant concept $c$, one for each domain constant $c$, where $c^\inter = \{ c \}$.
\end{itemize}
We iteratively construct the following compositional roles:
\begin{itemize}
  \item the universal role $\top$ where $\top^\inter = \Delta^\inter \times \Delta^\inter$,
  \item the role intersection $R \sqcap S$ where $(R \sqcap S)^\inter = R^\inter \cap S^\inter$,
  \item the role union $R \sqcup S$ where $(R \sqcup S)^\inter = R^\inter \cup S^\inter$,
  \item the role negation $\neg R$ where $(\neg R)^\inter = \top^\inter \setminus R^\inter$,
  \item the role inverse $R^{-1}$ where $(R^{-1})^\inter = \{ (b, a) \mid (a, b) \in R^\inter \}$,
  \item the role composition $R \circ S$ where $(R \circ S)^\inter = \{ (a, c) \mid (a, b) \in R^\inter \wedge (b, c) \in S^\inter \}$,
  \item the transitive closure $R^+$ where $(R^+)^\inter = \bigcup_{n \geq 1} (R^\inter)^n$,
  \item the transitive reflexive closure $R^\star$ where $(R^\star)^\inter = \bigcup_{n \geq 0} (R^\inter)^n$,
  \item the role restriction $R\vert_C$ where $(R\vert_C)^\inter = R^\inter \cap (\Delta^\inter \times C^\inter)$,
  \item the identify $\mi{id}(C)$ where $(\mi{id}(C))^\inter = \{ (a, a) \mid a \in C^\inter \}$.
\end{itemize}
The iterated composition $(R^\inter)^n$ is constructed inductively with $(R^\inter)^0 = \{ (a, a) \mid a \in \Delta^s \}$ and $(R^\inter)^{n+1} = (R^\inter)^n \circ R^\inter$.

The \emph{complexity} of a concept or role is the number of rules that are applied during its construction, or, equivalently, the size of its syntax tree.
We only consider a finite subset of roles and concepts up to complexity bound $c_\text{max}$.

\subsection{Features}%
\label{sub:Features}

Let $C, D$ be concepts and $R, S, T$ roles up to complexity $c_\text{max}$.
We construct the following Boolean features $f$ and define their values $f^\inter$ as follows:
\begin{itemize}
  \item empty feature $\mi{Empty}(C)$ where $(\mi{Empty}(C))^\inter = \top$ iff $C^\inter = \emptyset$,
  \item concept inclusion $C \sqsubseteq D$ where $(C \sqsubseteq D)^\inter = \top$ iff $C^\inter \subseteq D^\inter$,
  \item role inclusion $R \sqsubseteq S$ where $(R \sqsubseteq S)^\inter = \top$ iff $R^\inter \subseteq S^\inter$,
  \item nullary $\mi{Nullary}(p)$ where $(\mi{Nullary}(p))^\inter = \top$ iff $p$ is nullary state predicate and $p \in s$.
\end{itemize}

Similarly, we construct the following numerical features:
\begin{itemize}
  \item count $\mi{Count}(C)$ where $(\mi{Count}(C))^\inter = |C^\inter|$,
  \item concept distance $\dist(C, R, D)$ where $(\dist(C, R, D))^\inter$ is the smallest $n \in \mathbb{N}_0$ such that there are objects $o_0, \ldots, o_n$ with $o_0 \in C^\inter$, $o_n \in D^\inter$, and $(x_i, x_{i+1}) \in R^\inter$ for all $0 \leq i < n$.
    If $C^\inter$ is empty or no such $n$ exists, then $(\dist(C, R, D))^\inter = \infty$,
  \item the sum concept distance $\sumdist(C, R, D)$ where $(\sumdist(C, R, D))^\inter = \sum_{x \in C^\inter} \dist^\inter(\{ x \}, R, D)$,
  \item role distance $\rdist(R, S, T)$ where $(\rdist(R, S, T))^\inter$ is the smallest $n \in \mathbb{N}_0$ such that there are objects $a, o_0, \ldots, o_n$ with $(a, o_0) \in R^\inter$, $(a, o_n \in T^\inter$, and $(x_i, x_{i+1}) \in R^\inter$ for all $0 \leq i < n$.
    If $R^\inter$ is empty or no such $n$ exists, then $(\rdist(C, R, D))^\inter = \infty$,
  \item the sum role distance $\sumrdist(R, S, T)$ where $(\sumrdist(R, S, T))^\inter = \sum_{r \in R^\inter} \rdist^\inter(\{ r \}, S, T)$.
\end{itemize}

\section{Implementation}%
\label{sec:implementation}

We implemented the propositional theory $T(\mathcal{S}, \mathcal{F})$ as an \emph{Answer Set Program} (ASP)~\cite{lifschitzAnswerSetsLanguage2016} with \texttt{clingo}~\cite{gebserPotasscoPotsdamAnswer2011}.
We use the library \texttt{pddl}\footnote{\url{https://github.com/AI-Planning/pddl}} for PDDL parsing and \texttt{DLPlan} \cite{drexlerDLPlan2022} for feature generation in the same way as \cite{drexler:icaps2022,frances:aaai2021}.

\begin{algorithm}
  \textbf{Input:} Class $\mathcal{Q} = P_1, \ldots, P_k$, max complexity $c_\text{max}$
  \\
  \textbf{Output:} Generalized FOND policy $\pi$ for $\mathcal{Q}$
  \begin{algorithmic}
    \State $c_\text{min} \gets 1$; $T \gets \emptyset$, $\pi \gets (\emptyset, \emptyset, \emptyset)$; $S \gets \emptyset$; $U \gets \emptyset$
    \ForAll{$P \in \mathcal{Q}$}
      \If{\Call{CheckPolicy}{$\pi,P$}}
        \State Add $P$ to $S$
        \Continue
      \EndIf
      \State Add $P$ to $T$
      \State $\text{cost}_\text{max} \gets \infty$
      \ForAll{$c \in \{c_\text{min}, \ldots, c_\text{max} \}$}
      \State $\mathcal{F} \gets \text{\Call{GenerateFeatures}{$c$}}$
      \State $\pi_\text{new} \gets \text{\Call{Solve}{$T, \mathcal{F}, c, \text{cost}_\text{max}$}}$
      \If{$\pi_\text{new}$}
      \State Add $P$ to $S$
      \State $\pi \gets \pi_\text{new}$;
      $c_\text{min} \gets c$;
      $\text{cost}_\text{max} \gets \text{cost}(\pi_\text{new}) - 1$
      \EndIf
      \EndFor
    \EndFor
    \State \Return $\pi$
  \end{algorithmic}
  \caption{
    Incremental solver for learning a policy.
  }
  \label{alg:incremental-learning}
\end{algorithm}

The control loop is shown in \autoref{alg:incremental-learning}, which uses an iterative approach to solve a class of problems $\mathcal{Q}$.
Starting with the smallest instance (in terms of number of objects) of $\mathcal{Q}$, it iteratively tests whether the current policy solves the current problem $P$.
If the policy fails on $P$, then $P$ is added to the training set and a new policy for the complete training set is determined.
This approach avoids the need to select good training instances manually, as the solver determines which instances to use for training.

In each iteration, the solver increments the maximal complexity of all features in $\mathcal{F}$ from the complexity needed for the last policy up to a maximal complexity $c_\text{max}$.
After finding a policy $\pi$, it continues with features of higher complexity but with a total cost bounded by the cost of the last policy.
This way, the solver often finds an expensive first policy with low-complexity features and then iteratively improves this policy with features of higher cost while exploiting the upper bound on the total cost.
Without this optimization, the solver would often run out of memory for high-complexity features because the number of possible feature combinations grows too large.

For \Call{CheckPolicy}{$\pi, P$}, we simulate the policy $\pi$ on $P$, i.e., we choose each action according to $\pi$ and then randomly choose one outcome.
Note that a successful run does not guarantee that the policy solves $P$.
For this reason, we repeat each check ten times and assume that the policy is a solution if it succeeds every time.

\begin{listing*}
  \caption{
    The clingo code for selecting features and good transitions on a given set of features and a set of instances.
    For each instance, the input contains facts for states \lstinline|state| (partitioned into \lstinline|alive|, \lstinline|goal|, and implicit dead states), transitions \lstinline|trans|, features \lstinline|feature|, and feature evaluations \lstinline|eval|.
    The solver selects features \lstinline|selected| and good transitions \lstinline|good_trans| such that there is an outgoing good transition for each alive state and such that it can distinguish good and non-good transitions as well as alive, dead, and goal states with the selected features.
  }
  \label{lst:clingo}
\begin{minted}{clingo}
{ selected(F) } :- feature(F).
1 { good_trans(I, S1, S2) : trans(I, S1, A, S2), safe_action(I, S1, A) } :- alive(I, S1), not goal(I, S1).
{ good_trans(I, S1, S2) : trans(I, S1, A, S2), alive(I, S2) } :- alive(I, S1), not goal(I, S1).
:- alive(I, S), not good_trans(I, S, _), not goal(I, S).
:- good_trans(I, _, S), not alive(I, S).
bool_eval(I, S, F, 1) :- eval(I, S, F, V), V > 0.
bool_eval(I, S, F, 0) :- eval(I, S, F, V), V = 0.
bool_dist(I1, S1, I2, S2) :- state(I1, S1), state(I2, S2), bool_eval(I1, S1, F, V1), bool_eval(I2, S2, F, V2),
                             selected(F), V1 != V2.
trans_delta(I, S1, S2, F, -1) :- trans(I, S1, _, S2), eval(I, S1, F, V1), eval(I, S2, F, V2), V2 < V1.
trans_delta(I, S1, S2, F, 0)  :- trans(I, S1, _, S2), eval(I, S1, F, V),  eval(I, S2, F, V).
trans_delta(I, S1, S2, F, 1)  :- trans(I, S1, _, S2), eval(I, S1, F, V1), eval(I, S2, F, V2), V2 > V1.
trans_diff(F, I1, S11, S12, I2, S21, S22) :- trans_delta(I1, S11, S12, F, D1), trans_delta(I2, S21, S22, F, D2), D1 != D2.
distinguished(I1, S11, S12, I2, S21, S22) :- trans_diff(F, I1, S11, S12, I2, S21, S22), selected(F).
:- alive(I1, S11), alive(I2, S21), not bool_dist(I1, S11, I2, S21), good_trans(I1, S11, S12),
   trans(I2, S21, _, S22), not good_trans(I2, S21, S22), not distinguished(I1, S11, S12, I2, S21, S22).
:- state(I1, S1), state(I2, S2), not bool_dist(I1, S1, I2, S2), goal(I1, S1), not goal(I2, S2).
safe_state(I, S) :- goal(I, S).
safe_state(I, S1) :- alive(I, S1), safe_state(I, S2) : good_trans(I, S1, S2).
:- alive(I, S), not safe_state(I, S).
safe_action(I, S1, A) :- state(I, S1), trans(I, S1, A, _), alive(I, S2) : trans(I, S1, A, S2).
crit_state(I, S2) :- alive(I, S1), trans(I, S1, _, S2), not alive(I, S2).
:- alive(I1, S1), crit_state(I2, S2), not bool_dist(I1, S1, I2, S2).
#minimize { C,F : selected(F), feature_complexity(F, C) }.
#program limit_feature_cost(c).
:- #sum { C,F : selected(F), feature_complexity(F, C) } > c.
#program min_feature_complexity(c).
:- C < c : selected(F), feature_complexity(F, C).
\end{minted}
\end{listing*}

Each call to \Call{Solve}{$T, \mathcal{F}, c, \text{cost}_\text{max}$} instantiates the \texttt{clingo} program shown in \autoref{lst:clingo} with the training set $T$, features $\mathcal{F}$ up to complexity $c$, and an upper limit of the total cost of $\text{cost}_\text{max}$, implementing the theory $T(\mathcal{S}, \mathcal{F})$.
As an optimization, if it is known that there is no policy with features up to complexity $c-1$, then \lstinline|min_feature_complexity(c)| enforces the use of at least one feature of complexity $c$.
This helps the solver to quickly prune solution candidates that do not lead to a solution.
As further optimizations, instead of using the ranking $V(s, d)$ as introduced in Section~\ref{sec:learning}, we incrementally label all states where all selected transitions lead to the goal as \emph{safe} and require that all \emph{alive} states are also \emph{safe}.
Additionally, we do not aim to distinguish all dead states from alive states.
Instead, we only compare alive states to \emph{critical} states, which are those states that are dead-ends but have an incoming transition from an alive state.
Finally, we pre-process the state space $S$ by pruning all dead states that are not critical.

\section{Additional Correctness Results}
\subsection{Blocks}%
\label{ssub:blocks}

\newcommand*{\nclear}{\ensuremath{\mi{c}_\mi{on}}\xspace}
\newcommand*{\nmis}{\ensuremath{\mi{m}}\xspace}
\newcommand*{\non}{\ensuremath{\mi{on}}\xspace}

In \emph{blocks3ops}, the player can move a block from the table to a stack or vice versa and he may move a block between two stacks.
The learned policy $\pi_\text{blocks}$ uses three features:
\begin{enumerate*}
  \item a numerical feature $\nclear \equiv |\exists \mi{on}_G^+.\,\mi{clear}|$, the number of blocks which should be above another block that is currently clear,
  \item a numerical feature $\nmis \equiv |\mi{on} \setminus \mi{on}_G|$, the number of misplaced blocks, i.e., a block that is stacked on a block that it should not be stacked on,
  \item a numerical feature $\non \equiv |on|$, the number of currently stacked blocks.
\end{enumerate*}

Note that if $\nmis = 0$, then $\nclear$ is the number of incorrectly placed blocks: If a block $a$ is correctly placed, i.e., it is stacked on the correct block and all blocks below $a$ are also stacked correctly, then there cannot be any clear block that should be below $a$.
On the other hand, if $a$ is not correctly placed, then there must be some clear block $b$ that should be below it, as otherwise there is block $c$ on $b$ that is misplaced, and so $m > 0$.

The rules of the learned policy $\pi_\text{blocks} = \pi_{R,B}$ are the following:
\begin{alignat*}{3}
  &r_1: \quad &\{ \nclear > 0, \nmis = 0 \} &\mapsto \{ \inc{\non}, \dec{\nclear} \} \mid \{ \dec{\non} \}
  \\
  &r_2: \quad &\{ \nmis > 0, \non > 0 \} &\mapsto \{ \dec{\nmis}, \dec{\non} \} \mid \{ \dec{\nmis}, \dec{\non}, \inc{\nclear} \} \mid
  \\ &&&\qquad \{ \dec{\non} \} \mid \{ \dec{\non}, \inc{\nclear} \}
\end{alignat*}
The learned policy does not contain any state constraints as there are no dead-ends in blocks.
\begin{propositionE}
  The general policy $\pi_\text{blocks} = \pi_{R,D}$ solves the class  $\mathcal{Q}_\text{blocks3ops}$ of solvable FOND \emph{blocks3ops} problems.
\end{propositionE}
\begin{proofE}
  Let $P$ be any instance of $\mathcal{Q}_\text{blocks3ops}$ and $\pi_P$ the corresponding concrete policy.
  First, as there are no dead-ends in \emph{blocks3ops}, any \emph{blocsk3ops} policy is dead-end-free.

  Next, we show that $\pi_P$ is complete.
  First, assume $m > 0$, and so there are is at least one tower.
  It is easy to see that rule $r_2$ contains all the combinations how unstacking a block from the tower may affect the features: \non is always decreased, while \nmis may be either decreased or remain unchanged and \nclear may be either increased or remain unchanged.
  Second, assume $m = 0$.
  We may already have a tower segment, where the lowest block should be stacked on another block, and hence the segment needs to be unstacked.
  This will be done by decreasing \non while leaving the other features unchanged.
  Otherwise, there is already a partially correct tower (consisting of 1 block or more) and the action compatible with $\{ \inc{\non}, \dec{\nclear} \}$ puts the correct block on the partial tower until the tower is completed.

  Finally, the policy $\pi_P$ is descending over tuple  $t = \la \nmis \non, \nclear, \non \ra$.
  Rule $r_2$ always decreases \non and decreases \nmis or leaves it unchanged and hence always decreases $t$.
  If the condition of rule $r_1$ is satisfied, then $\nmis = 0$ and so $\nmis \non = 0$.
  If rule $r_1$ is applied, either $\nclear$ is decreased (while increasing $\non$), or $\nclear$ remains unchanged and $\non$ is decreased.
  In both cases, $t$ decreases.
  By \autoref{thm:descending-policies}, $\pi_P$ is a solution for $P$ and so $\pi_\text{blocks}$ is a solution for $\mathcal{Q}_\text{blocks3ops}$.
\end{proofE}

\section{Proofs}%
\label{sec:Proofs}
\printProofs
}{}

\end{document}